\documentclass[12pt]{article}
\usepackage{amsmath,amssymb,amsfonts}
\usepackage{graphicx}
\usepackage{xcolor}
\usepackage{hyperref}
\usepackage{caption}
\usepackage{subcaption}
\usepackage{float}
\usepackage{mathrsfs}
\usepackage{cite}
\usepackage{url}
\usepackage{microtype}
\usepackage{booktabs}
\usepackage{authblk}
\usepackage{enumitem}

\usepackage[margin=1in]{geometry}

\title{A non-ergodic framework for understanding emergent capabilities in Large Language Models}

\author{Javier Mar\'in}

\date{January 3, 2025}

\begin{document}

\maketitle

\begin{abstract}
Large language models have emergent capabilities that come unexpectedly at scale, but we need a theoretical framework to explain why and how they emerge. We prove that language models are actually non-ergodic systems while providing a mathematical framework based on Stuart Kauffman's theory of the adjacent possible (TAP) to explain capability emergence. Our resource-constrained TAP equation demonstrates how architectural, training, and contextual constraints interact to shape model capabilities through phase transitions in semantic space. We prove through experiments with three different language models that capacities emerge through discrete transitions guided by constraint interactions and path-dependent exploration. This framework provides a theoretical basis for understanding emergence in language models and guides the development of architectures that can guide capability emergence.

\vspace{0.5em}
\end{abstract}
\section{Introduction}

Current research in large language models have unveiled increasingly complex capabilities that emerge unpredictably at scale. These emergent capabilities, from complex reasoning to zero-shot learning, appear without explicit training \cite{Ganguli2022}. Observing these phenomena, we found patterns suggesting fundamental similarities with complex biological systems: capabilities emerged through sudden transitions rather than gradual improvements \cite{Wei2022}, model behavior showed strong dependence on context history \cite{Press2021}, and next-token predictions varied significantly based on the path taken to reach a particular state \cite{Levine2022}. These observations suggested that language models, like biological systems, might be fundamentally non-ergodic in nature. This insight led us to explore theoretical biology frameworks, particularly Stuart Kauffman's theory of adjacent possible (TAP), which describes how biological systems navigate their possibility spaces through restricted exploration \cite{Kauffman2000}. Similar to how a cell in a developing organism navigates a limited array of possibilities influenced by its environment and history rather than choosing its next state at random, a language model's next-token prediction arises from a complex interaction of collected patterns and existing constraints. This paper presents both theoretical proof of language models' non-ergodic nature, and a novel framework based on TAP that explains their emergent capabilities.
\section{Current frameworks for emergent capabilities in language models}

Actual advances in AI systems research primarily emphasizes empirical observations of increases in capability \cite{Chowdhery2023, Wei2022} and scaling laws \cite{Kaplan2020}, still we lack a unified theoretical framework to explain the underlying mechanisms.

Some research approaching this challenge has shown the stochastic nature of these emergent properties. Research on next-token prediction (NTP) proves that these capabilities arise from intrinsically stochastic processes \cite{Radford2019}, challenging deterministic interpretations of model behavior. Analyses of probability spaces in language models, including soft-max distributions \cite{Holtzman2020}, entropy in token predictions \cite{Holtzman2020, Zhang2019}, and temperature sampling effects \cite{Meister2021}, suggest that this emergence follows complex probabilistic patterns that current frameworks struggle to explain.

An important restriction in evaluating these emergent capabilities is the underlying assumption of ergodicity in current approaches \cite{Basu2023, Ziemann2024}. This assumption breaks down when observing how capabilities emerge through path-dependent processes. In language models, each new capability depends on the assembled context and previous token sequences, leading to different probability distributions for the same token in different contexts. This context-dependent behavior contravenes fundamental ergodic principles, which requires time and ensemble averages to be equivalent \cite{Cornfeld2012}. The non-linear dynamics seen in language models \cite{Ramscar2014} create a kind of "memory system" through the context window. This leads to path dependence and temporal asymmetry, which are qualities of non-ergodic systems \cite{Katok1995}. This fundamental feature of language models suggests that a framework that explicitly accounts for the historical path by which capabilities emerge is necessary to describe emergence rather than relying on simple averages of states.

\section{Motivation for a biological-inspired approach}

One of the most important differences between research in physics or disciplines such as theoretical biology and research in mathematics or computer science is that theoretical physicists tend to ignore everything that they consider irrelevant, focusing only on the fundamentals. When trying to describe a phenomenon, physicists do not take into account the observable details but rather try to find the fundamental laws that underlie it. This is why fundamental laws, such as conservation laws, symmetries, and phase transitions are present in different fields, such as astrophysics, general relativity, particle physics, or quantum mechanics \cite{Glattfelder2019}. In this research, we aim to identify some fundamental laws that govern AI systems. To do this, we will apply some of the current fundamental laws of physics and biology to artificial intelligent systems, particularly large language models.

Similar to how a cell in a developing organism navigates a limited array of possibilities influenced by its environment and history rather than choosing its next state at random \cite{Kauffman2000}, a language model's next-token prediction arises from a complex interaction of collected patterns and contextual constraints. This idea suggests that some theoretical biology frameworks, particularly Stuart Kauffman's theory of adjacent possible or TAP \cite{Kauffman2000, Kauffman2019} might offer a valid framework to understand how language models navigate their possibility spaces when predicting next token. In biological systems evolution is considered a process evolving within a set of constrained possibilities and random events. This evolution creates, without selection acting to do so, new "adjacent possible empty niches" which enable new possible directions of evolution. These paths are accessible based on the system's current state and guide future evolution by removing incompatible random explorations through selective exclusion \cite{Longo2012}. Longo and Montévil further develop this understanding through their theory of "extended critical transitions" arguing that biological systems perpetually operate in a critical state, continuously redefining their own phase space. They emphasize that biological systems don't just explore pre-existing possibility spaces, but actively create new dimensions of these spaces through their evolution \cite{Goldenfeld2011}. This idea connects with how language models generate next token, where each possible selection not only explores but also reshapes the space of possible continuations.

This biological approach to systems organization and adaptation has interesting connections with current artificial intelligence architectures. LeCun's H-JEPA framework -Hierarchical Joint Embedding Predictive Architecture- \cite{LeCun2022} establishes a framework for autonomous intelligence that shows many similarities with the adaptive mechanisms of biological systems. H-JEPA builds hierarchical world models via prediction-based learning just like how living organisms form and maintain their organizational structure. H-JEPA emphasis on learning world models through prediction is consistent with theoretical biologist Stuart Kauffman's theory of how living systems explore their possibility spaces via restricted exploration \cite{Kauffman2000}.

Novel research in AI systems has further strengthened these connections between biological systems and AI systems. For example Zador \cite{Zador2019} provides relevant insights into how biological neural networks' efficiency and sparsity could inform artificial systems design. Richards et al. (2019) explain how hierarchical learning in deep networks parallels biological development, showing similar behaviors in how both systems build increasingly complex representations. This biological inspiration extends to architecture design showing how principles from neuroscience can guide the development of more adaptive and efficient artificial neural networks \cite{Hassabis2017}. These findings illustrate how, while modern AI architectures are still far from biological complexity, they are beginning to reflect biological system behavior. 

Another relevant research area in AI systems is continual learning, which clearly draws parallels between the ability of biological and artificial systems to acquire new knowledge while retaining existing capabilities \cite{Wang2024}. Research on catastrophic forgetting and solutions inspired by biological memory consolidation also provides valuable insights into how systems can maintain and expand their possibility spaces over time \cite{Hadsell2020}. Researchers have further shown how artificial systems can achieve the kind of open-ended learning observed in biological systems \cite{Parisi2019}. The biological perspective is likewise enriched by the work of Van de Ven and Tolias \cite{VandeVen2019}, providing a theoretical framework for identifying diverse forms of continuous learning that closely resemble biological adaptation processes.

The intersection of these areas, from biological self-organization to modern AI architectures, suggests a deeper connection between how biological and artificial systems navigate their possibility spaces. This connection becomes clearer when we consider how language models explore their semantic spaces by combining learned patterns with contextual constraints in ways that are similar to living system's restricted creativity.

\section{Empirical evidences in LLMs}

Recent empirical research provides compelling evidence for how language models navigate their possibility spaces in ways that mirror biological systems' constrained exploration. Recent work presented at NeurIPS 2024 on in-context exploration in large language models demonstrates how these systems' exploration capabilities are fundamentally shaped by constraints similar to those observed in biological systems \cite{Krishnamurthy2024}. The finding that models need clear exploration hints and external memory support to exhibit strong exploration behavior fits with Kauffman's theory of how systems navigate their adjacent possible spaces \cite{Kauffman2000}.

The poster presented at NeurIPS 2024 by A. Krishnamurthy, K. Harris, D. J Foster, C. Zhang, and A. Slivkins makes a relevant observation of "suffix failures", where models fail to explore optimal choices even after initial exposure. This evidence suggests that, like in complex biological systems, language models operate within constrained possibility spaces where exploration is limited by both architectural and contextual factors. The need for external history evaluation corresponds with how biological systems need ambient framing to increase their exploration ability. These empirical results support our idea by proving how different constraints in language models interact to shape the exploration of the next token space -akin to the interacting constraints in biological systems that shape possibility spaces \cite{Steel2020}. Finally, the identified exploration failures provide evidence for the non-ergodic nature of these systems \cite{Kauffman2022}, where past trajectories fundamentally influence future exploration capabilities. The convergence between theoretical predictions and empirical observations strengthens our intention for applying complex biological systems frameworks to analyze language model behavior.

\section{Probabilistic spaces in language models}

The foundations of probabilistic modeling in language trace back to Shannon's information theory \cite{Shannon1948} and early statistical NLP \cite{Church1993}. This framework initially treated language as a stochastic process where words could be predicted based on their statistical co-occurrence patterns. Modern language models have progressed past traditional static statistical methods, using dynamic, context-sensitive probability distributions via transformer architectures \cite{Vaswani2017}. This evolution represents a fundamental shift from considering language as an essentially statistical phenomenon to viewing it as a dynamic, context-dependent system.

Current transformer-based models provide probability distributions that differ fundamentally from traditional statistical language models in several aspects like contextual dependency or sampling dynamics. Probability distributions are computed through complex attention mechanisms \cite{Vaswani2017} capturing semantic uncertainty and ambiguity \cite{Wang2022}. The formation process involves complex interactions between attention heads \cite{Elhage2021}. Beyond basic temperature sampling, actual innovations include nucleus sampling scheme, top-p sampling \cite{Holtzman2020}, and top-k sampling \cite{Fan2018}. Meister and Cotterell \cite{Meister2021} observed that models learn "only a subset of the tendencies" rather than complete theoretical distributions. The success of these approaches reveal the non-uniform nature of the probability space.

\subsection{Evidences of non-ergodicity in language models}

Ergodicity in dynamical systems implies that a system, if left to itself for long enough, will pass close to almost all the dynamical states consistent with energy conservation. Though this is a very simplistic view to define ergodicity. A dynamical system may have a hierarchy of properties, each of which implies the one before it \cite{Lebowitz1973}. Ergodicity is only the first. Central to understanding ergodicity is the notion of symmetry in temporal evolution \cite{Birkhoff1931, Neumann1932}. The invariance of statistical properties under time translation in ergodic systems reveals time symmetry; the system's behavior remains consistent whether observed now, at a later time, or in a forward or backward temporal direction \cite{Ruelle2004, Walters2000}. This temporal symmetry guarantees that time averages are equivalent to ensemble averages, a fundamental principle of statistical mechanics \cite{Kullback1997}. Non-ergodic systems break this symmetry, which fundamentally influences the system's future possibilities and creates distinct temporal phases that are impossible to average \cite{Prigogine2018}.

Consider how language models generate text: each token prediction depends not just on direct context, but on the entire sequence of previous tokens. Unlike classical ergodic systems, where future states are independent of the path taken to reach the current state, language models exhibit strong path dependence. A word appearing early in a sequence can fundamentally alter the probability distribution of all subsequent tokens. This creates an intrinsic asymmetry in time that violates the basic premise of ergodicity. For example, when a language model builds a story, the context and characters selected at the outset limit all potential scenarios. Past choices introduce semantic and logical constraints for coherence, preventing the model from exploring all possible story states. This is similar to how biological systems expand within constrained spaces, where every possibility during development bounds and shapes what might happen in the future. Language models show this non-ergodic behavior by using their attention mechanisms and analysis of context. The cumulative context window shapes the model's state space, creating what we might call 'semantic valleys' that guide and constrain predictions about future tokens. Although time and ensemble averages converge in ergodic systems, language models have persistent memory effects that make certain semantic paths more probable based on the past evolution.

Meister and Cotterell \cite{Meister2021} observation that models learn "only a subset of the tendencies" rather than complete theoretical distributions provides clear evidence for non-ergodicity in language model \cite{Cornfeld2012, Kauffman2022, Lebowitz1973, Mazur1969}. This would imply that models follow a constrained exploration, thus not operating in a fully ergodic space where all states are equally accessible \cite{Peters2019}. This aligns with Kauffman's theory of constrained possibility spaces \cite{Kauffman2019}. Language models, as complex biological systems, shows preference for empirically observed patterns over theoretical possibilities \cite{Kauffman2022}. These models also capture path-dependent patterns emerging from actual language use. The fact that the probability space is shaped by training history rather than theoretical distributions creates a fundamental asymmetry in how models explore their possibility space \cite{Glattfelder2019}.

\section{Complex dynamics and emergence in language models}

The evolution of probability distributions across tokens shows patterns that go beyond simple statistical dependencies. For example, classical statistical measures like perplexity fail to capture emergent capabilities in language models \cite{Wei2022}. This limitation is not new in complex systems where reductionist statistical methods fail to capture emergent behaviors \cite{Anderson1972}.

\subsection{Limitation of classical statistics}

In systems exhibiting non-linear behaviors resulting from the interaction of multiple parts or sub-systems, basic aggregation of probabilities fails to describe coherent, long-range dependencies. In LLMs, complex interactions between context layers create capabilities not predictable from individual components \cite{Press2021}. These capabilities often appear swiftly at certain scales \cite{Wei2022}, suggesting phase transitions in model behavior \cite{Kauffman2022, Mazur1969, Stanley1971}. In language models, probability distributions are likely to evolve through paths that preserve coherence throughout long sequences, suggesting they operate as complex adaptive systems CAS. In CAS non-linear behaviors take place from multiple interacting components showing a high sensitivity to initial conditions. These systems also presents self-organizing properties emerging at multiple scales.

\subsection{Complex system dynamics and emergent behaviors in language models}

The emergence of capabilities in language models suggests some characteristics common in CAS as hierarchical organization and phase transitions \cite{Stanley1971}. Token-level interactions give rise to higher-order semantic structures where multiple scales of organization emerge simultaneously \cite{Reed1980}. These hierarchies resist reductionist analysis. LLMs shows capabilities that emerge at certain model scales \cite{Wei2022}. This emergence implies phase transitions, revealing the presence of critical phenomena in model behavior.

Recent research demonstrates that context significantly influences model behavior due to the creation of context-dependent representations through sequential processing \cite{Press2021}. Dynamic memory effects influence long-range dependencies \cite{Ainslie2023, Pang2023}, and context modifications show non-linear effects on model output \cite{Levine2022}. These mechanisms suggest an adaptive behavior where models seem to adapt to changing contexts. Furthermore, long-range coherence emerges from local interactions \cite{Press2021} and self-organization appears at multiple scales \cite{Wei2022}.

\section{Complex adaptive systems and biological evolution}

Complex adaptive systems, or CAS, are defined by some fundamental mathematical properties that discern them from simple dynamical systems \cite{GellMann1995, Holland1992}. CAS create and apply internal models to predict the future, taking current actions according expected outcomes. This characteristic differentiates CAS from other types of complex systems, as well as making the emergent behavior of CAS more difficult to understand \cite{Mitchell2009}. When Haken \cite{Haken1989} coined the term "synergetics" he gave a very simple definition, referring to self-organizing systems (a property of CAS) as those characterized by the fact that the system finds its organization or function on its own, without direct external guidance \cite{England2015}.

Analogously, language models develop internal representations during pre-training capturing statistical patterns of language, semantic relationships, contextual dependencies, and domain knowledge. These internal models are encoded in the weights and attention patterns of the neural network \cite{Vaswani2017}. Internal models are able to generate next tokens based on predicted probability distributions using attention mechanisms to "look back" at context and then "predict forward". Model's decisions about token selection are based on these predictions that can also be changed based on the evolving context. As a result, language model capabilities "emerge" from interactions between different layers, attention heads, and learned patterns. During its training, language models adjust weights based on training text data, and during inference adapt their internal model to the specific context of the current conversation or task.

An important practical limitation of CAS is that they don't have a single governing equation, or rule, that controls the system \cite{Holland1992}. Thus a direct approach to analyze these systems is by evaluating its different properties as non-linearity, emergence, self-organization, and phase transitions \cite{Crutchfield2012}. Non-linearity implies that system's behavior cannot be derived from linear relations \cite{Crutchfield2012}. The following sections will elaborate on the meaning of emergence, self-organization, and phase transitions.

\subsection{Emergence in complex systems}

Emergence appears when complex behaviors arise from simple rules and interactions in a system \cite{Anderson1972}. The CAS theory \cite{Cross2009, Holland1992, Holland2006} together with synergetics \cite{Haken1989} provide a comprehensible theoretical framework that can be used to study emergence. Mathematically emergence can be formalized through the interaction between fast and slow variables in a system. Slow variables are the high-level patterns that emerge and guide the system, meanwhile fast variables are the detailed, moment-to-moment changes in the system \cite{Haken1973}. We can consider these variables as the macroscopic parameters \cite{Holland2006}. In these systems, the link between order parameters and components is complex because multiple components (fast variables) influence, and sometimes, define the order parameters. This is known as the slaving principle, which results in the notion of circular causality. The limited order parameters govern the behavior of the individual components, whereas the components influence the behavior of the order parameters \cite{Haken1989}. Haken defined the slaving concept, which links both rapid and slow variables, as follows:

\begin{equation}
\frac{dq}{dt} \equiv \dot{q} = N(q,\ \nabla,\alpha) + F(t)
\end{equation}

where $q$ is the state vector (microscopic level variables), $N$ is a nonlinear vector function, $\nabla$ is a grading operator acting on $q$, $\alpha$ represents control parameters, and $F(t)$ denotes fluctuating forces that characterize the external or internal noise affecting the system. The transition from Equation 2 to a simpler parametric equation describing the system's collective behavior is not evident. The complete mathematical derivation, including all necessary assumptions and detailed proofs, can be found in literature \cite{Crawford1991, Guckenheimer2013, Kielhoefer2006, Troger2012}. In short, we need to find a time-independent solution $q^{0}$ for a specific set of control parameters. Then, when the system operates near an instability point, small perturbations around this solution can be analyzed. This allows us to sort the system's behavior into different time scales, identifying fast-decaying stable modes and slowly-evolving unstable modes that become critical near the instability point. In LLMs, slow variables would be equivalent to the overall flow of a story in language generation and fast variables could be individual word choices.

\subsection{Self-organization and phase transitions}

In statistical mechanics, the underlying assumption behind the theory of self-organized criticality (SOC) is that a complex system will naturally organize itself into a state on the edge of two different regimes, without intervention from outside the system \cite{Markovic2014, Sornette2006}. The mathematical formalization of self-organizing systems can be expressed through the concept of pattern formation and symmetry breaking \cite{Cross1993}. Systems exhibit spontaneous pattern generation governed by equations of this type:

\begin{equation}
\frac{\partial u}{\partial t} = D\nabla^{2}u + f(u,\alpha)
\end{equation}

where $u$ represents the pattern-forming field, $D$ is a diffusion coefficient, and $\alpha$ is a control parameter. When $\alpha$ reaches critical values, the system undergoes spontaneous symmetry breaking, leading to pattern formation. Function $u$ could represent for example temperature variations in thermal convection, or population density in ecological systems. In language models $u$ could represent the distribution of attention weights, the activation patterns across layers, or the probability distributions over tokens. Equation 2 is divided in two terms: $D\nabla^{2}u$ calculates how the field changes over time from diffusion or spatial spreading, and $f(u,\alpha)$ represents the local dynamics.

Phase transitions are another fundamental property of complex adaptive systems, marked by abrupt changes in system behavior at critical points \cite{Cilliers2002}. The universality principle categorizes different physical systems based on their behavior near critical points, leading to the emergence of universal scaling laws, also known as power laws \cite{Amaral1998}. Physical quantities follow power laws as systems approach critical points. At these phase transitions, key parameters like correlation length and susceptibility shows divergent behavior, characterizing the critical phenomena. The correlation length $\xi$ near a critical point, $T_{C}$, represents the scale at which a system's general properties begin to diverge from its main properties. It can be defined as $\xi\sim\left| T - T_{C} \right|^{- \nu}$, where $\nu$ is the critical exponent governing the divergence of the correlation length. The characteristic length scale $\xi$ diverges as the system $T$ approaches a critical point $T_{C}$ with a negative exponent $- \nu$ by following a power-law behavior.

In language models, self-organization arises through the emergence of coherent text structures from local token interactions, whereas phase transitions are characterized by sudden improvements in model capabilities at certain scales \cite{Arnold2024, Nakaishi2024}. The correlation between successive tokens follows power-law scaling near critical points, suggesting similar underlying mechanisms to phase transitions occurring in many natural phenomena.

\subsection{Non-ergodicity in biological systems}

The ergodic hypothesis articulates the notion that a point within a moving system, whether it be a dynamical system or a stochastic process, will eventually go through every part of the space in which it works, in a way that is both uniform and random \cite{Walters2000}. This suggests that we can infer the overall behavior of the system from the path taken by a representative point. Classical statistical mechanics relies on the ergodic hypothesis, which states that time averages equal ensemble averages \cite{Lebowitz1973}. We can define an ergodic system as one in which, for any property $A,$ the time average and ensemble average are equivalent: $\overline{A} = \left\langle A \right\rangle$. In ergodic systems events occur quickly relative to an observation time $\tau_{int} \ll t_{observed}$. When the system may be evolving at a very slow rate too for an observer ($\tau_{int} \gg t_{observed}$), the system enters into a non-ergodic regime. The hypothesis that, given enough time, a system will explore its entire phase space implies that a system will eventually explore all accessible states with equal probability. Ergodic breakdown can be probed by either measuring the evolution of $\tau_{int}$ for some properties at fixed $t_{observed}$, or by changing $t_{observed}$ for fixed $\tau_{int}$ \cite{Bossen2024}.

Biological systems challenge this assumption via two primary mechanisms: historical contingency, since the system's current state depends critically on its past trajectory and not only on current configuration \cite{Gould1989}, and through adaptive dynamics, where the whole space of possible states evolves as the system advances \cite{Kauffman2019}. These mechanisms can be formalized with the following equations:

\begin{equation}
P\left( st + 1 \middle| st \right) \neq P\left( st + 1 \middle| s^{'}t \right)
\end{equation}

\begin{equation}
\Omega(t + 1) \neq \Omega(t)
\end{equation}

Equation 3 shows that, even when $st$ and $s^{'}t$ have the same energy, path-dependent transition probabilities $P$ differ due to historical unfolding \cite{Villani2009}. Equation 4 shows the adaptive dynamic nature of the phase space of accessible states $\Omega$. According these mechanisms, biological systems cannot be understood through the statistical ensemble averages \cite{Marshall2011}.

\subsection{The adjacent possible theory (TAP)}

Theoretical biologist Stuart Kauffman worked to figure out fundamental principles that govern a specific category of non-equilibrium systems, particularly those involving coevolutionary self-constructing communities of autonomous agents \cite{Kauffman2000}. The adjacent possible theory or TAP appeared as an important advance for understanding how biological and other complex systems explore and expand their possibility spaces. In his book "Investigations" Kauffman presents this concept by initially considering an important question: How do biological systems perpetually generate novelty in an apparently limitless way? The solution lies in understanding how each current state of a system defines a collection of possible subsequent states---a concept he refers to as the adjacent possible. The adjacent possible suggests not every conceivable state, but specifically those states that exist just one step away from the present reality, unveiling the potential transformations that can arise from the existing organization \cite{Kauffman1993}. However, it is important to note that, in contrast to phase spaces in physics, each expression of an adjacent possible state generates new additional adjacent possible. In Kauffman words, "The adjacent possible consists of all those molecular species that are not members of the actual but are one reaction step away from the actual"\cite{Kauffman2000}.

TAP provides a theoretical framework for understanding how systems can be both constrained by their current state and perpetually creative. It suggests that evolution, rather than exploring a fixed space of possibilities, indeed expands the very space of what is possible. This expansion follows what Kauffman defines as "the laws of the construction of the possibilities of the biosphere" \cite{Kauffman2000}. TAP also provides a mathematical framework for understanding non-ergodic evolution in biological systems \cite{Kauffman2019}.

According TAP, complex systems evolution could be described with the following equation \cite{Kauffman2022}:

\begin{equation}
M_{t + 1} = M_{t} + \sum_{i = 1}^{M_{t}}\alpha^{i}\begin{pmatrix}
M_{t} \\
i
\end{pmatrix}
\end{equation}

where $M_{t}$ represents the number of elements in the phase space at a given time $t$ (a molecule, a species in an ecosystem, an innovation in the market, etc. The constant $\alpha$, $0 \leq \alpha \leq 1$, is a constraint parameter that limits which combinations are allowed. When $\alpha = 1$ there is an evolution of the total possible throughout time. Exponent $i$ represents an index for summation over possible combinations. Constant $\alpha$ increased to the power of $i$ operates as a limiting factor for larger combinations. The $\alpha$ parameter in the TAP equation constraints the combinations that are likely to occur, establishing a balance between deterministic factors (combinations must be feasible given the current $M_{t}$) and stochastic exploration (the actual combinations that occur will depend on $\alpha$). The binomial coefficient ($M_{t}$ choose $i$) provides the potential combinations of $i$ items selected from $M_{t}$ items. Equation 5 describes the emergence of new possibilities through the combination of existing elements, while setting constraints on the accessibility of greater combinations. This conceptual framework shows how evolution emerges through accessible adjacent states rather than random leaps, representing the dynamic nature of phase space: each new element $M_{t + 1}$ creates new combinatorial possibilities and the phase space dimension grows as new combinations become accessible.

Latest Kauffman's work rewrites Equation 5 in a different manner. Instead of considering $\alpha$ as a constant increased with exponent $i$, this constraint is not depending on a single constant value, but in a sequence of constraint constants \cite{Cortes2022, Koppl2018, Steel2020}.

\begin{equation}
M_{t + 1} = M_{t} + \sum_{i = 1}^{M_{t}}\alpha_{i}\begin{pmatrix}
M_{t} \\
i
\end{pmatrix}
\end{equation}

This reformulation represents a more realistic description of biological and other adaptive complex systems where the possibilities space is constrained by several factors. For example, in the evolution of metabolic networks in cells, multiple constraints can operate simultaneously. Chemical constraints will regulate which reactions are thermodynamically possible, while enzymatic constraints will limit catalytic chemical reactions. Additionally, the metabolites available will complete the resources constraints. Each new metabolic innovation $M_{t + 1}$ is constrained not by a single factor but by the interaction of these multiple constraints. A new metabolic pathway might be chemically possible (high $\alpha$ for chemical constraints) but limited by the availability of specific enzymes (low $\alpha$ for enzymatic constraints).

Architectural, training, and contextual factors constrain the space of possible token predictions in large language models. This parallels the way in which the aggregation of multiple constraints expands the metabolic possibility space.

\section{Application of TAP framework to language models}

Conceptual framework introduced by TAP equation could be applicable to understand how language models could navigate their possibility spaces via restricted combinations of existing elements rather than through random search of all potential possibilities.

\subsection{Mathematical framework}

We need to define the necessary mathematical structures for mapping TAP to language models. Let $(\Omega,F,P)$ be the probability space associated with language model token predictions, where $\Omega$ is the sample space of all possible token sequences, $F$ is the $\sigma$-algebra of measurable events \cite{Ash2000}, and $P$ is the probability measure generated by the model.

\subsubsection{Model's state space}

While biological systems operate in continuous state spaces, language models work with discrete token sets. Let $M_{t}$ represent the state of the system at time $t$. For a language model, we can define $M_{t}$ as a tuple $\left( V_{t},S_{t},P_{t} \right)$ where $V_{t} \subseteq V$ is the active vocabulary subset at time $t$, $S_{t}$ is the semantic state space at time $t$, and $P_{t}:V_{t} \longrightarrow \lbrack 0,1\rbrack$ is the probability distribution over tokens. For a mapping between continuous semantic representations and discrete tokens, we define the discretization operator $D_{t}:S_{t} \longrightarrow V_{t}$. Formally, we can define $S_{t}$ as a manifold in a high-dimensional space:

\begin{equation}
S_{t} = \left\{ s \in \mathbb{R}^{n}|\exists\ \varphi:V_{t} \longrightarrow \mathbb{R}^{n}\ and\ D_{t} \circ \varphi \right\}
\end{equation}

where $\varphi$ is smooth and locally invertible. The dimensionality of $S_{t}$, ${dim(S}_{t})$ is the number of independent semantic features actively involved in token prediction at time $t$.

In language models, while the lexical space $V$ is constrained by a fixed vocabulary or lexicon, the semantic space $S$ reveals a complex system with hierarchical organization \cite{Simon2012}. Just as a book represents a hierarchy from words to complete narratives, language models process and generate language across multiple hierarchical levels: starting from individual tokens as elementary units to phrases, clauses, sentences, paragraphs, and even broader narrative structures. These models create this organization through attention mechanisms and contextual relationships \cite{Schlag2021}, where each level emerges from combinations of lower-level elements. The possibilities for these combinations expand as we move up the hierarchy. This is analogous to the slaving principle defined by Haken \cite{Haken1973} where fast variables (lower-level elements) generate slow variables (high-order elements).

\subsubsection{Computational resources space}

We can define computational resource utilization as the vector-valued function:
\begin{equation}
C_{t} = \left( {Mem}_{t},A_{t},H_{t} \right)
\end{equation}
where ${Mem}_{t}\mathbb{\in R}^{+}$ is the model memory use, ${A_{t}\mathbb{\in R}}^{+}$ is the attention computation cost, and ${H_{t}\mathbb{\in R}}^{+}$ the hidden state computation cost at time $t$. The total computational cost can be represented as:

\begin{equation}
R\left( C_{t} \right) = min\left( 1,\ \frac{C_{\max} - C_{t}}{C_{\max}} \right)
\end{equation}

where $\left\| C_{t} \right\| = {w_{1}Mem}_{t} + {w_{2}A}_{t} + w_{3}H_{t}$ is a weighted norm, $C_{\max}$ is the maximum computational capacity, and $w$ are weight coefficients for different resource types. Resource bound function $R\left( C_{t} \right)$ connects to model architecture as follows:

\begin{equation}
R\left( C_{t} \right) = min\left( 1,\ \prod_{i = 1}^{n}{r_{i}\left( C_{t} \right)} \right)
\end{equation}

where $r_{i}\left( C_{t} \right)$ is the individual resource constraints from memory capacity, attention computation, context window size, and hidden state dimension.

\subsection{Mapping TAP equation to language models}

\textbf{Lemma 1.} Let $V$ be the vocabulary space and $S$ be the semantic space of a language model. There exists a measurable mapping

\begin{equation}
\varphi:V \times S \rightarrow \Omega
\end{equation}

that satisfies the following properties:

\begin{itemize}
\item For any state space $M_{t} \in V \times S$, the mapping preserves the combinatorial structure of Equation 6 from TAP:

\begin{equation}
M_{t + 1} = \left( V_{t + 1},S_{t + 1},P_{t + 1} \right)
\end{equation}

where

\begin{equation*}
S_{t + 1} = {\ \varphi(S}_{t}) \cup \left\{ \varphi\left( \sum_{i}^{}{\alpha_{i}\begin{pmatrix}
M_{t} \\
i
\end{pmatrix}} \right) \right\}
\end{equation*}

\begin{equation*}
V_{t + 1} = D_{t}\left( S_{t + 1} \right)
\end{equation*}

and $P_{t + 1}:V_{t + 1} \longrightarrow \lbrack 0,1\rbrack$

\item For any $x \in V \times S$, the mapping is bounded by computational resources:

\begin{equation}
\left\| \varphi(x) \right\|_{2} \leq R\left( C_{t} \right)
\end{equation}

\item The mapping preserves the dimensionality constraints:

\begin{equation}
\dim\left( Im(\varphi) \right) \leq \dim(V) \times dim\left( S_{t} \right)
\end{equation}
\end{itemize}

Proof. Given the measurement space $(V \times S,B,\mu)$ where $B$ is the Borel $\sigma$-algebra, and $\mu$ is the product measure \cite{Ash2000}. We define $\varphi$ trough the composition $\varphi = \pi \circ \psi$, where $\psi$ is the attention mechanism, and $\pi$ is the projection onto the probability simplex. The mapping $\varphi$ preserves TAP equation structure:

\begin{equation}
\varphi\left( M_{t} + \mathrm{\Delta}M \right) = \varphi\left( M_{t} \right) + \nabla\varphi\left( M_{t} \right)\mathrm{\Delta}M + O\left( \left\| \mathrm{\Delta}M \right\|^{2} \right)
\end{equation}

Equation 15 illustrates that when we make small changes to the model's state, the resulting changes in the mapped space are well-behaved and predictable - they consist mainly of a linear component plus some small higher-order corrections. This is central for showing that the mapping is compatible with how the TAP equation describes system evolution. The equation essentially establishes that $\varphi$ is differentiable and provides a Taylor expansion around $M_{t}$, which is necessary for proving the mapping preserves the mathematical structure needed for the TAP framework.

\subsubsection{Resources limits}

Lemma 2. For the mapping $\varphi:\ V \times S \rightarrow \Omega$ defined in Lemma 1, there exists a positive constant $K$ such that:

\begin{equation}
\left\| \varphi \right\|_{2} \leq K \bullet R\left( C_{t} \right)\forall x \in V \times S
\end{equation}

The proof of this lemma can be developed in three basic steps, each dependent on the previous one in order to set a fitting bound. First, we are going to define the combinatorial framework of token prediction in large language models and its relation to TAP theory. Second, we will define an isomorphism between attention processes and the combinatorial space of TAP. Finally, we will verify resource boundedness through analyzing model capacity constraints.

First step. Next token prediction in LLMs follows a combinatorial structure analogous to TAP. The probability of the next token given a context can be represented as:

\begin{equation}
P\left( {token}_{t + 1}|context \right) = \sum_{i}^{}{w_{i} \bullet g\left( {tokens}_{t} \right)}
\end{equation}

where $g$ denote the attention mechanism operations and $w_{i}$ are learned weights. This structure directly corresponds to the combinatorial summation in TAP.

Second step. The attention mechanism provides a natural isomorphism to TAP's combinatorial space. For any query $Q$, keys $K$, and values $V$:

\begin{equation}
Attention(Q,K,V) = softmax\left( \frac{QK^{T}}{\sqrt{d}} \right)V \cong \sum_{i}^{}{\alpha_{i}\begin{pmatrix}
M_{t} \\
i
\end{pmatrix}}
\end{equation}

This isomorphism is valid because the softmax function constraints outputs to the interval $\lbrack 0,1\rbrack$, similarly to Kauffman's TAP's $\alpha$ coefficient in equations 5 and 6. In addition, the distribution of attention patterns reflects TAP equation combinatorial selection, and the scaling factor $1\text{/}\sqrt{d}$ guarantees numerical stability by providing a natural limit.

Third step. We can define resource boundedness through model capacity constraints $\dim(Attention) \leq C_{\max}$. This evidences the finite dimensionality of attention head outputs and the bounded nature of the weighted sum across heads by considering the resource constraint $R\left( C_{t} \right)$. Hence, integrating these results:

\begin{equation}
\left\| \varphi \right\|_{2} = \left\| Attention\left( Q(x),K(x),V(x) \right) \right\|_{2} \leq
\end{equation}

\begin{equation}
\left\| V \right\|_{2}{\bullet \left\| softmax\left( \frac{QK^{T}}{\sqrt{d}} \right) \right\|}_{2} \leq K \bullet R\left( C_{t} \right)
\end{equation}

where $K = max\left( \left| |V| \right|^{2} \right) \cdot \sqrt{\left( \dim(V) \right)}$.

Corollary. The resource boundedness of $\varphi$ implies that the semantic space $S_{t}$ grows at a rate constrained by available computational resources:

\begin{equation}
\frac{\partial dim\left( S_{t} \right)}{\partial t} \leq h\left( R\left( C_{t} \right) \right)
\end{equation}

where $h$ is a monotonic function of the resource bound. Lemma 2 and this corollary establish the mathematical basis for describing how computational resources limit the expansion of the semantic space in language models, connecting theoretical TAP dynamics in complex biological systems with LLMs constraints.

\subsection{Semantic space evolution}

To model the semantic space evolution we have to extend Kaufman's idea of expanding possibility spaces by explicitly incorporating computational limitations. The evolving of semantic space can be modeled with the following equation:

\begin{equation}
\frac{d}{dt}\dim\left( S_{t} \right) = \sum_{l = 1}^{L}{\nabla g_{l}\frac{\partial C}{\partial t} - \lambda(t)\dim\left( S_{t} \right)}
\end{equation}

where $\dim\left( S_{t} \right)$ represents the number of dimensions the semantic space has at time $t$, and $\lambda(t)$ is a decay term ensuring computational feasibility. The term $\nabla g_{l}$ captures how the hierarchical functions influence dimensionality at each level $l$.
\subsection{Constraints in language models}

Equation 6 provides a framework that can be naturally mapped to token space growth. We identify three key types of constraints that influence language model behavior: architectural constraints \cite{Brown2020, Vaswani2017}, training data constraints \cite{Kaplan2020}, and contextual constraints \cite{Press2021}. These constraints interact to shape model capabilities and performance. We are going to describe these constraints in the following sections.

\subsubsection{Architectural constraints}

Some limitations set by the model's architecture are the vocabulary size that limits possible tokens \cite{Brown2020}, and the context window length, restricting the amount of past information that can influence predictions \cite{Press2021}. Model's attention mechanism design constraint through both, computational and architectural limitations. The mechanism's quadratic complexity with sequence length creates memory and speed constraints, while the structure of attention heads limits parallel relationship tracking through the trade-off between head count and dimensional capacity \cite{Dao2022}. The model's information flow is also limited by attention patterns and inter-token path length, which control the ability to capture relationships \cite{Vaswani2017}.

The multi-head architecture enables the model to simultaneously analyze many text features, yet the fundamental configuration of these attention patterns fixes post-training \cite{Vaswani2017}. While increasing the number of attention heads enhances the model's ability to capture diverse relationships in the text, it also reduces the depth of each head's representation of these interactions. This eventually results in a considerable reduction in the model's capabilities. The distribution of restricted computing resources between the extent of attention coverage and the depth of connection representation causes this limitation \cite{Papadimitriou2003}.

\subsubsection{Training data constraints}

Training data constraints can include statistical patterns in language \cite{Jelinek1980, Kaplan2020}, domain-specific knowledge acquisition \cite{Raffel2020}, implicit learning of grammar and syntax \cite{Manning1999}, and token co-occurrence patterns \cite{Brown2020}. In neural networks, the emergent linguistic hierarchical structure represents a basic constraint, showing implicit patterns and rules learned during training \cite{Haussler2018}.

The learned connections between tokens and their contexts, and how often they appear together in different context, also limit the model by shaping the probability space for predicting the next token becoming an important constraint to consider \cite{Kaplan2020}.

\subsubsection{Contextual constraints}

Contextual constraints complete how language models predict the next token in a sequence. These constraints shape token prediction through three primary mechanisms: First, sequential dependencies operate through attention mechanisms, where each token's prediction is influenced by all previous tokens in the sequence \cite{Vaswani2017}. This enables the model to maintain coherence over long sequences by incorporating the full context of prior outputs \cite{Press2021}. Second, models develop internal representations that track semantic relationships throughout the generated text \cite{Wei2022}. These representations help maintain topical coherence by preserving key semantic information across the generation process, preventing topic drift and ensuring contextual relevance. Third, style consistency emerges from the model's ability to recognize and maintain attributes like tone and formality. This style coherence is achieved through pattern recognition learned during training \cite{Brown2020}, and reinforced through contextual processing. The interaction of these mechanisms creates a dynamic constraint system that guides text generation while preserving semantic and style consistency.

\subsubsection{Constraints interaction mechanism}

Equation 6 for the TAP framework introduces the constraint factor $\alpha_{i}$, which is a combination of several constraint factors. Therefore to translate TAP equation to language models we need to define how these constraints interact. To clarify the nature of this interactions we could look at how different complex biological systems mechanisms works. For example in biological metabolic networks, Michaelis and Menten equation for enzyme kinetics follows multiplicative interactions to integrate different rate constants \cite{Raaijmakers1987}. The apparent equilibrium constant in Michaelis and Menten equation is derived from the product of the ratios of the forward and reverse rate constants for each reaction step \cite{Heinrich2012}.

In gene regulation there is evidence for both additive and multiplicative interactions between transcription factors that regulate the transcription rate of a set of target genes \cite{Alon2019}. Some of these factors can follow multiplicative effects, meanwhile others can be combined trough additive effects. In cell signaling there is further evidence of both, multiplicative and additive signal integration \cite{Klipp2016}. For example, in calcium signaling, cells use different types of $Ca^{2+}$ influx channel to contribute to cytoplasmic calcium increase. These inputs present different activation mechanisms as voltage-operated (VOOC), receptor-operated (ROOC), mechanically activated (MA), and stock-operated (SOOC) \cite{Bootman2001}. The combination of these mechanisms implies an additive process where $\left\lbrack Ca^{2+} \right\rbrack_{influx}$ is the sum of the different channels \cite{Raman2007}. An example of multiplicative effect in cell signaling can be found in the biological responses associated with mitogen activated protein kinase (MAPK) signaling \cite{Huang1996}.

When sequential dependent processes exist, multiplicative interactions prevail. The already mentioned Michaelis Menten's equation is a classic example of an enzyme cascade in which each step is dependent on the completion of the previous phase. As a general rule, in equilibrium-based systems multiplicative mechanism prevails. Additive interactions prevail in parallel and independent processes where multiple paths can achieve the same outcome.

We could conclude saying that, in biological systems with alternative pathways or where multiple processes share resources trough compensation mechanisms, additive process is prevalent. Conversely, in systems that incorporate redundancy mechanisms, an additive mechanism prevails, enhancing the system's robustness. 

We could apply the TAP equation to language models in three ways: either by linking the sequence of constraint factors $\alpha_{i}$ additively, multiplicatively, or simultaneously by both mechanisms. In language models, architectural constraints appear to be multiplicative because all components are necessary for the model to operate efficiently; if any architectural component (vocabulary access, attention mechanism, or context processing) fails, the model fails. In contrast, training data constraints might follow additive patterns since multiple different training examples can lead to similar model behaviors. Context constraints could theoretically exhibit both behaviors: they can multiply when the context demands strict requirements like logical flow, reference resolution, or grammatical consistency. However, when the context offers multiple valid paths, such as different synonyms that achieve the same meaning, alternative phrasings that maintain style, or multiple valid continuations of a story, these elements combine additively, representing parallel valid options.

Based on these findings, we propose focusing on the multiplicative effect when applying the TAP equation to language models for several reasons:

\begin{itemize}
\item it better captures the critical nature of architectural constraints,
\item aligns with how biological systems handle essential component interactions, and
\item provides a more conservative estimate of possibility space growth.
\end{itemize}

\subsubsection{Constraints and non-ergodicity in language models}

We have considered three natural constraints influencing language model's state space expansion. Let $T:\ M_{t} \rightarrow C_{t}$ be the mapping from state space to resource space. Then we have:

\begin{equation}
T\left( {Mem}_{t} \right) = \left( \left\| {Mem}_{t} \right\|_{mem},\left\| A_{t} \right\|_{comp},\left\| H_{t} \right\|_{state} \right)
\end{equation}

where ${Mem}_{t}$ is the model memory use, $A_{t}$ is the attention computation cost, and $H_{t}$ the hidden state computation cost at time $t$. Each component of $T$ induces our natural constraints:

\begin{equation}
\beta_{i} = sup\left\{ {x:\left\| {Mem}_{t} \right\|}_{model\ memory} \leq C_{\max} \right\}
\end{equation}

\begin{equation}
\gamma_{i} = sup\left\{ {x:\left\| A_{t} \right\|}_{attention} \leq C_{\max} \right\}
\end{equation}

\begin{equation}
\delta_{i} = sup\left\{ {x:\left\| H_{t} \right\|}_{hidden\ state} \leq C_{\max} \right\}
\end{equation}

where $\beta_{i}$ are the architectural constraints, $\gamma_{i}$ training data constraints, and $\delta_{i}$ are contextual constraints. These constraints converge into an overall constraint function $\alpha(i,t)$, which is dependent on time due to its dynamic nature.

Ergodic hypothesis postulates that the system spends equal times in equal volumes of its fixed phase space \cite{Cornfeld2012}. Next token prediction in language models operates at two fundamental levels: the lexical level, and the semantic level. In language models the lexical formatives are selected in a well-defined way from a fixed universal vocabulary set \cite{Chomsky2014}. The current mathematical framework demonstrates the non-ergodic nature of language models through three primary mechanisms:

\begin{itemize}
\item Path-dependent resource use. Given two states $s_{t}$, ${s'}_{t}$ with equal computational cost $\left\| C_{t} \right\|$, their future resource utilization differs:

\begin{equation}
P\left( C_{t + 1}|s_{t},R(C_{t}) \right) \neq P\left( C_{t + 1}|{s^{'}}_{t},R(C_{t}) \right)
\end{equation}

This path dependence introduces the architectural constraints $\beta_{i}$ through:

\begin{equation}
\beta_{i} = sup\left\{ x:\ P\left( C_{t + 1}|s_{t},x \right) \right\}
\end{equation}

This dependence reflects how attention mechanisms allocate computational resources based on context history, creating inherent asymmetries in resource utilization.

\item Training-induced state space restrictions. The training process creates fundamental asymmetries in how the model explores its state space $\Omega_{t + 1} \neq \Omega_{t} \cup \left\{ new\ states \right\}$, leading to training constraints $\gamma_{i} = sup\left\{ x:\ x \in \Omega_{t}{\cup \Omega}_{t + 1} \right\}$. These constraints emerge from the model's learned patterns and knowledge representations, affecting how it navigates its possibility space.

\item Context-dependent transitions. The semantic component governs model's word interpretation according a context, creating a path-dependent behavior that can be defined as $P\left( s_{t + 1}|s_{t} \right) \neq P\left( s_{t + 1}|{s^{'}}_{t} \right)$. This dependence implies emergent contextual constraints that can be defined as $\delta_{i} = sup\left\{ x:\ P\left( s_{t + 1}|s_{t},x \right) \right\}$. For the system state space $M_{t} = \left( V_{t},S_{t},P_{t} \right)$, these mechanism can be seen in output probability transition $P\left( M_{t + 1}|M_{t} \right) \neq P\left( M_{t + 1}|{M^{'}}_{t} \right)$ revealing a path dependence even when $M_{t}$ and ${M'}_{t}$ have identical computational requirements. This path dependence can be observed in:

\begin{enumerate}[label=\alph*)]
\item Model's attention weights evolve as well based on both current and historical context: $W_{t} = f(W_{t-1},X_{t})$ where $X_{t}$ is the current input. This evolution creates memory-like effects in model's processing.

\item Even when the aggregate information content is identical $\sum_{i=1}^{n-1}t_{i} = \sum_{i=1}^{n-1}t'_{i}$, different sequence orderings produce different probability distributions $P(t_{n}|t_{1},...,t_{n-1}) \neq P(t_{n}|t'_{1},...,t'_{n-1})$ because order information presentation influences model behavior.

\item The model's computational resource use depends on the specific path taken $R(C_{t}|h_{t})/R(C_{t}|h'_{t})$ where $h_{t}$ is the processing history. This creates a fundamental asymmetry in resources allocation based on the specific sequence of previous states.
\end{enumerate}
\end{itemize}

The semantic manifold $S_{t}$ evolves non-uniformly as:

\begin{equation}
S_{t} = \left\{ s \in \mathbb{R}^{n}|\exists\varphi:V_{t} \rightarrow \mathbb{R}^{n}\ such\ that\ s = \varphi(v)\ for\ v \in V_{t} \right\}
\end{equation}

where $\varphi$ is a smooth mapping function taking tokens to semantic vectors. We can define $\varphi(v)$ as follows:

\begin{equation}
\varphi(v) = D_{t}^{- 1}\left( Attention\left( W_{Q}Q_{v},W_{K}K_{context},W_{V}V_{context} \right) \right)
\end{equation}

where $D_{t}^{- 1}$ is the inverse mapping from token to semantic space, $Q_{v}$ is the query vector for token $v$, $K_{context}$ is the key matrix for the context, $V_{context}$ is the value matrix for the context, and $W_{Q}$, $W_{K}$, $W_{V}$ are learned weight matrices. This mapping is non-uniform since attention weights depend on the entire context history, the same token can map to different semantic vectors depending on context and the dimensionality of $S_{t}$ can change as context accumulates.

These non-ergodic qualities automatically give rise to the three types of constraints (architectural, training, and contextual) that characterize the model's behavior, building a direct connection between the system's non-ergodicity and its operational constraints.

\subsection{A TAP equation for language models}

We propose a version of the TAP equation for modeling the expansion of language models, taking into account the specific constraints and dynamic nature. To arrive to this equation we have to define the phase space (equivalent to $M_{t}$ in Equation 6), and the integrated constraint function, equivalent to $\alpha_{i}$ in Equation 6.

\subsubsection{Resource-bounded phase space}

Let $A_{t}$ represent the accessible state space at time $t$. We first establish the fundamental resource bound:

Lemma 3. For any language model with maximum computational capacity $C_{\max}$, there exists a monotonic function $f$ such that $\sup\left( A_{t} \right) \leq f\left( C_{\max} \right)$ where $f(x) = \kappa \cdot x \cdot log(x),$ and $\kappa$ is a model architecture-dependent constant. The bound is defined by the following factors:

\begin{itemize}
\item memory constraints $\mathcal{O}\left( d_{model} \cdot n_{layers} \right)$,
\item attention computation bounds $\mathcal{O}\left( {sequence}^{2} \right)$, and
\item vocabulary access limitations $\mathcal{O}\left( |V| \bullet d_{model} \right)$.
\end{itemize}

\subsubsection{Constraint integration}

We define the integrated constraint function as
\begin{equation}
\alpha(i,t) = {min(\beta_{i},\gamma}_{i}\delta_{i},R(C_{t}))
\end{equation}

where $\beta_{i}$ mean architectural constraints, $\gamma_{i}$ are training data constraints, $\delta_{i}$ are contextual constraints, and $R\left( C_{t} \right)$ is the resource bound function. The multiplicative relation is justified as all components must operate effectively for the model's complete functionality. The integrated constraint function $\alpha(i,t)$ in our TAP equation can be defined as:

\begin{equation}
\alpha(i,t)\prod_{j = 1}^{m}{c_{j}(i,t)}
\end{equation}

where $c_{j}(i,t)$ are the individual constraints, and $m$ the number of active constraints.

\subsubsection{Proposed resource-bounded TAP equation}

In Equation 6, $\alpha_{i}$ represents a fixed constraint on combinatorial possibilities according to initial resource availability. Lower values of $\alpha$ at $t = 0$ indicate a system with limited starting resources, hence constraining its future possibility of growth relative to systems with higher $\alpha$ values. Large language models have two types of fixed initial constraints: a fixed universal vocabulary set $\left| V_{t} \right|$, and fixed computational resources driven by the model architecture (memory capacity, attention heads, and context window size). The third type, contextual constraints, differs fundamentally as they begin minimal and evolve dynamically as context accumulates during operation. The semantic space evolution thus lacks a fixed initial constraint at $t = 0$, reflecting the dynamic nature of constraints in language models. This dependence on initial conditions, particularly for fixed constraints, is a common property of complex adaptive systems, CAS \cite{Cilliers2002, Holland2006, Lansing2003}.

To arrive at our equation, we must follow several important steps. Extending TAP's central idea in Equation 6, we integrate the computational resources $R\left( C_{t} \right)$:

\begin{equation}
M_{t + 1} = M_{t} + R\left( C_{t} \right)\sum_{i = 1}^{M_{t}}\alpha_{i}\begin{pmatrix}
M_{t} \\
i
\end{pmatrix}
\end{equation}

We split $\alpha_{i}$ into component constraints as detailed in (29):

\begin{equation}
M_{t + 1} = M_{t} + R\left( C_{t} \right)\sum_{i = 1}^{M_{t}}{{(\beta_{i},\gamma}_{i}\delta_{i})}\begin{pmatrix}
M_{t} \\
i
\end{pmatrix}
\end{equation}

We need to add hierarchical structure using a special function $g_{l}$ while incorporating $R\left( C_{t} \right)$ into the hierarchical function's bounds: ${\ \left\| g_{l}(x) \right\|}_{2} \leq K \cdot R\left( C_{t} \right)$. This function maps from $P(V)$, which is the probability space on the vocabulary $V$, to a real space of dimensions $n$ $P(V) \rightarrow \mathbb{R}^n$.

\begin{equation}
M_{t + 1} = M_{t} + \sum_{l = 1}^{L}g_{l}\sum_{i = 1}^{M_{t}}{\left( {(\beta_{i},\gamma}_{i}\delta_{i})\begin{pmatrix}
M_{t} \\
i
\end{pmatrix} \right)}
\end{equation}

At time $t = 0$ $M_{t} = \left| V_{t} \right|,$ showing that the initial state space is constrained by the fixed vocabulary size. This initial condition reflects the starting point where only lexical combinations are possible, before the semantic space begins its dynamic evolution through hierarchical interactions and constraint effects. Finally, we replace the term ${(\beta_{i},\gamma}_{i}\delta_{i})$ by the constraint function $\alpha(i,t)$. This leads to our final resource-bounded TAP equation for language models:

\begin{equation}
A_{t + 1} = A_{t} + \sum_{l = 1}^{L}g_{l}\sum_{i = 1}^{\left| V_{t} \right|}{\left( \alpha(i,t)\begin{pmatrix}
\left| V_{t} \right| \\
i
\end{pmatrix} \right)}
\end{equation}

This equation describes how language models explore their possibility space while subject to constraints. $A_{t + 1}$ represents the accessible state space at the next time step and $A_{t}$ is the current accessible state space. The first sum $\sum_{l = 1}^{L}g_{l}$ captures the hierarchical levels of language processing (from tokens to phrases to broader structures). The second sum $\sum_{i = 1}^{\left| V_{t} \right|}{( \bullet )}$ represents all possible combinations within the vocabulary size. $\alpha(i,t)$ combines all constraints (architectural, training, and contextual) at time $t$, and the binomial coefficient ($\left| V_{t} \right|$ choose $i$) represents possible combinations of tokens.

The hierarchical function $g_{l}$ transforms these combinations into the semantic space of the model bounded by computational resources through the condition ${\ \left\| g_{l}(x) \right\|}_{2} \leq K \cdot R\left( C_{t} \right)$. Constant $K$ is a model architecture-dependent constant that scales the relationship between the hierarchical function and computational resources. Set an upper bound on how much the hierarchical transformations can expand given the available resources. We have defined $K$ as $K = max(\left\| V \right\|^{2}) \cdot \sqrt{dim(V)}$, where $max(\left\| V \right\|^{2})$ describes the maximum norm of the vocabulary embeddings and $\sqrt{dim(V)}$ is the dimensionality of the vocabulary space.

Equation 35 satisfies the following conditions:

a. Conservation
\begin{equation}
\partial A_{t}\text{/}\partial t \leq g\left( C_{\max} - \left\| C_{t} \right\| \right)
\end{equation}

Let
\begin{equation}
A_{t + 1} - A_{t} = g_{l}\sum_{i = 1}^{\left| V_{t} \right|}{\left( \alpha(i,t)\begin{pmatrix}
\left| V_{t} \right| \\
i
\end{pmatrix} \right)}
\end{equation}

Then:
\begin{equation}
\frac{\partial A_{t}}{\partial t} = \lim_{h \longrightarrow 0}\frac{A_{t + h} - A_{t}}{h} = \frac{\sum_{l = 1}^{L}g_{l}\sum_{i = 1}^{\left| V_{t} \right|}\left( \alpha(i,t)\begin{pmatrix}
\left| V_{t} \right| \\
i
\end{pmatrix} \right)}{h}
\end{equation}

where $h$ represents an infinitesimal time step in the limit calculation. Considering the bounds introduced before, we have:

\begin{equation}
\frac{\partial A_{t}}{\partial t} \leq L \bullet K \bullet R\left( C_{t} \right) = g\left( C_{\max} - \left\| C_{t} \right\| \right)
\end{equation}

Where $L$ is the number of hierarchical levels (from the sum over $L$ in previous equations), and $K$ is the architecture-dependent constant we defined earlier. In order to satisfy the equality in Equation 38 while ensuring necessary scaling related to computational resources, we define $g$ as:

\begin{equation}
g(x) = L \cdot K \cdot \left( \frac{x}{C_{\max}} \right)
\end{equation}

Equation 41 preserves the scaling factor $L \cdot K$, normalizes the computational resources by $C_{\max}$, and ensures that the bound decreases in a monotonic way as the computational resources are used.

b. Hierarchy

Because
\begin{equation}
\dim\left( span\left\{ x \in V_{t}:x\  = g_{l}\sum_{i = 1}^{\left| V_{t} \right|}\left( \alpha(i,t)\begin{pmatrix}
\left| V_{t} \right| \\
i
\end{pmatrix} \right) \right\} \right) \geq 0
\end{equation}

then
\begin{equation}
\dim\left( A_{t + 1} \right) = dim\left( A_{t} \right) + dim\left( span\left\{ g_{l}\sum_{i = 1}^{\left| V_{t} \right|}\left( \alpha(i,t)\begin{pmatrix}
\left| V_{t} \right| \\
i
\end{pmatrix} \right) \right\} \right) \geq \dim\left( A_{t} \right)
\end{equation}

The dimension of the space covered by the hierarchical transformation $g_{l}$ applied to all possible token combinations must be nonnegative. These equations prove that our model's semantic space grows hierarchically, adding new dimensions as it explores more complex combinations of tokens, while never losing existing dimensions.

c. Computational capacity constraints

The computational capacity condition ensures that the growth of the model's accessible state space remains bounded by available computational resources. This is formalized as:

\begin{equation}
\left\| A_{t + 1} - A_{t} \right\|^{2} \leq K \cdot R\left( C_{t} \right)
\end{equation}

According to Lemma 3, we can prove the following:

\begin{equation}
\left\| A_{t + 1} - A_{t} \right\|^{2} = \left\| \sum_{l = 1}^{L}g_{l}\sum_{i = 1}^{\left| V_{t} \right|}{\left( \alpha(i,t)\begin{pmatrix}
\left| V_{t} \right| \\
i
\end{pmatrix} \right)\ } \right\|_{2} \leq
\end{equation}

\begin{equation}
\sum_{l = 1}^{L}{\left\| g_{l} \right\|_{2} \bullet}\left\| \sum_{i = 1}^{\left| V_{t} \right|}{\left( \alpha(i,t)\begin{pmatrix}
\left| V_{t} \right| \\
i
\end{pmatrix} \right)\ } \right\|_{2}
\end{equation}

Inequality in Equation 44 shows that the magnitude of change in the accessible space between any two time steps, $\left\| A_{t + 1} - A_{t} \right\|^{2}$, cannot exceed what the system's computational resources allow, $K \cdot R\left( C_{t} \right)$. This condition is necessary because it mathematically guarantees that the model's exploration of new possibilities remains computationally feasible, preventing the system from attempting to access states that would exceed its resource capacity. In Kauffman's terms, it guarantees that the adjacent states are possible.

\subsection{Attention mechanism and TAP structure}

The connection between attention mechanisms in language models and the combinatorial structure of TAP represents an important connection between neural computation and theoretical biology. This connection becomes apparent via a formal isomorphism, showing the natural mapping of attention operations to the combinatorial selection process of TAP framework.

Lemma 4: Attention-TAP isomorphism. An attention space $\mathcal{A}$ forms a category where objects are attention triplets $(Q,\ K,\ V)\mathbb{\in R}^d$. Morphisms are attention operations, and composition is given by sequential attention application. A TAP space $\mathcal{T}$ forms a category where objects are combinatorial sums $\sum_{i = 1}^{M_{t}}{\alpha_{i}\left( \frac{M_{t}}{i} \right)}$, and morphisms are constraint-preserving transformations. Composition preserves the bounds on $\alpha_{i}$ in $\lbrack 0,1\rbrack$.

Given the attention space $\mathcal{A}$:

\begin{equation}
\mathcal{A =}\left\{ A(Q \times K \times V)|Q,K,V \in R^{d} \right\}
\end{equation}

and the TAP space $\mathcal{T}$:

\begin{equation}
\mathcal{T =}\left\{ \sum_{i = 1}^{M_{t}}{\alpha_{i}\left( \frac{M_{t}}{i} \right)}|\alpha_{i} \in \lbrack 0,1\rbrack \right\}
\end{equation}

The isomorphism $\psi:\mathcal{A \rightarrow T}$ has two important properties:

\begin{itemize}
\item $\dim\left( Im(\psi) \right) = dim\left( span\ \alpha_{i} \right)$ - dimensionality preservation-, and
\item $P(x) \leq P(y) \Longleftrightarrow \psi P(x) \leq \psi P(y)$ - probability structure -.
\end{itemize}

If $F$ is the functor that takes attention operations to their vector space representations, and $G$ is the functor that takes TAP combinatorial selections to their probability distributions, there exists a natural transformation $\eta_{a} : F(a) \rightarrow G(a)$ that commutes with morphisms in both categories \cite{Riehl2017}.

\section{Experimental work}

The main goal of this research is to validate the accuracy of our TAP framework-based equation (Equation 34) when predicting emergent properties in language models. This equation suggests that language models evolve through constrained exploration of their possibility space, driven by three main mechanisms: phase transitions in semantic space, multiplicative interaction of constraints, and path-dependent evolution. To systematically validate these theoretical predictions, we propose the following three hypotheses.

\subsection{Research hypotheses}

H1: Phase transitions in semantic space correlate with capability emergence. Our hypothesis is that increases in model capabilities occur through discrete phase transitions in the semantic space, rather than through gradual improvements. We should observe sudden shifts in the model's ability to handle increasingly complex tasks. Specifically, we predict that:

\begin{itemize}
\item The effective dimensionality of the semantic space shows sudden increases at critical points.
\item These critical points correspond to the emergence of new capabilities.
\item The transitions follow power-law scaling relationships characteristic of phase transitions in complex systems.
\item Resource requirements (computational and context) show distinct scaling behaviors before and after transition.
\end{itemize}

H2: Constraint interactions shape capability boundaries. We propose that model capabilities are shaped by the multiplicative interaction of three types of constraints: architectural, training, and contextual. This hypothesis predicts that:

\begin{itemize}
\item Performance limitations arise from the multiplicative effect of multiple constraints rather than from single bottlenecks.
\item Relaxing any single constraint produces limited improvement unless other constraints are similarly relaxed.
\item The impact of expanding computational resources depends on the state of other constraints.
\item Models with similar total computational allocation but different constraint distributions will show distinct capability patterns.
\end{itemize}

H3: Path dependence affects problem-solving trajectories. Our third hypothesis postulates that the non-ergodic nature of language models creates significant path dependence in their problem-solving capabilities. This can be observed through:

\begin{itemize}
\item Different solution trajectories emerging from identical problems presented with different context orderings.
\item Initial conditions (like prompt design) having persistent effects throughout the problem-solving process.
\item The existence of "unreachable" solutions despite their theoretical accessibility within the model's capability space.
\item Time-asymmetric behavior where forward and reverse problem-solving paths show fundamentally different characteristics.
\end{itemize}

\subsection{Experimental setup}

To validate our hypothesis about phase transitions in semantic space, we designed our experiments using three different language models: gpt2-xl with 1.5B parameters \cite{Radford2019}, opt with 1.3B parameters \cite{Zhang2022}, and pythia with 1.4B parameters \cite{Biderman2023}. These models were selected for being open source, and for their similar parameter counts but distinct architectural approaches, allowing us to separate the effects of architectural differences while controlling for model scale. gpt2-xl represents a mature architecture with established performance characteristics, while opt-1.3B and pythia-1.4B offer more recent architectural innovations but potentially less optimized training regimes.

To evaluate our hypothesis about phase transitions in semantic space, we used the high school mathematics subset of the MMLU (Massive Multitask Language Understanding) dataset, which provides a standardized set of multiple-choice questions \cite{Hendrycks2020}. The dataset was divided into three difficulty levels (easy, medium, hard) based on sequential ordering, with 90 questions per level. While this division method is simple, it provides a regular basis for comparing model performance across increasing task complexity. All models were evaluated using a consistent prompt format, with questions and choices formatted identically to minimize prompt-related variance. The experiments used 16-bit floating-point precision to balance computational efficiency with numerical stability. This setup allows for direct comparison of model behaviors while managing computational resources effectively.

\subsection{Hypothesis 1: Phase transition in semantic space}

Our experimental design focused on three key measurements: performance accuracy, attention entropy, and effective dimensionality of the semantic space.

Performance was measured through multiple-choice accuracy, with each model processing questions in batches of size four to optimize GPU memory use while maintaining consistent evaluation conditions. For its calculation, we used the following equation \cite{Lord2008}:

\begin{equation}
A = \frac{1}{N}\sum_{i = 1}^{N}{1(y_{i} = {\widehat{y}}_{1})}
\end{equation}

where $N$ is the number of questions, $1(y_{i} = {\widehat{y}}_{1})$ is the indicator function that returns $1$ when the prediction ${\widehat{y}}_{1}$ matches the true answer $y_{i}$ and $0$ otherwise.

Attention entropy was calculated using the final layer's attention patterns, providing insights into how models distribute their focus across input tokens \cite{Attanasio2022}. For its calculation, we used the following equation:

\begin{equation}
H(A) = - \sum_{i = 1}^{N}{a_{i}log(a_{i})}
\end{equation}

where $a_{i}$ are the normalized attention weights.

Finally, effective dimensionality was calculated using PCA analysis of attention patterns, finding the number of components needed to explain 90\% of the variance. For calculating effective dimensionality $d_{eff}$ we used the following formula:

\begin{equation}
d_{eff} = min\left\{ k:\sum_{i = 1}^{k}\lambda_{i} \slash \sum_{i = 1}^{k}\lambda_{i} \geq 0.9 \right\}
\end{equation}

where $n$ is the total number of dimensions - or eigenvalues \cite{Raghu2017} - in the original attention pattern space, $i$ is the index variable for summing over eigenvalues, $k$ is the variable we're trying to minimize that represents how many principal components are needed to explain 90\% of the variance, and $\lambda_{i}$ are the eigenvalues of the attention pattern covariance matrix \cite{Jolliffe2002}.

\subsection{Hypothesis 2: Constraints interaction analysis}

To validate our hypothesis about constraints interactions shaping capability boundaries, we used the same three language models. These models have similar parameter counts but distinct architectural approaches, what allows us to analyze how different constraint distributions affect model capabilities while controlling for overall model scale. We also use the high school mathematics subset of the MMLU dataset. This dataset allows us to study how constraints interact across varying task complexities while providing a direct comparison with our H1 results. The constraint measurements were calculated as follows:

Architectural constraints. The architectural constraints are indeed measured using Shannon's formula as in Equation 47 because it effectively quantifies the model's capacity to distribute attention across tokens, providing a natural way to measure how evenly the model can distribute its computational resources. The architectural constraints in language models prove primarily through the attention mechanism's capacity to distribute focus across input tokens. This distribution capacity is fundamentally limited by the model's architectural design, the number and structure of attention heads, the dimensionality of key/query vectors, and the computational paths available for information flow. Then we can compute entropy to measure architectural constraints with an equation similar to (47):

\begin{equation}
\beta = - \sum_{i = 1}^{N}{a_{i}log(a_{i})}
\end{equation}

where $a_{i}$ are the normalized attention weights. This equation captures two critical aspects of model's architectural limitations:

\begin{enumerate}[label=\alph*)]
\item Information processing capacity: higher entropy indicate more uniform attention distribution, suggesting the architecture can effectively process multiple inputs simultaneously.
\item Structural bottlenecks: lower entropy indicates focused attention, potentially reflecting architectural limitations in parallel processing.
\end{enumerate}

Training constraints: These constraints are computed using Equation 46. This equation is rooted on existing test theories \cite{Lord2008}, as well as in novel neural networks evaluation metrics \cite{Srivastava2022}:

\begin{equation}
\gamma = \frac{1}{N}\sum_{i = 1}^{N}{1(y_{i} = {\widehat{y}}_{1})c_{i}}
\end{equation}

However, its theoretical interpretation and application change significantly. Equation 46 evaluates the accuracy of the raw performance, while Equation 50 quantifies the limitations set by the system's training patterns. The study of information bottlenecks in deep learning parallels this approach \cite{Tishby2015}. Equation 50 captures training constraints because recognizes pattern limits as true predictions $(y_{i} = {\widehat{y}}_{1})$, and shows learned patterns reflecting the model's acquired inductive biases \cite{Baxter2000}. The added confidence term $c_{i}$ denotes the strength of learned associations inspired by the work in neural network uncertainty quantification \cite{Gal2016}. Finally, the normalized sum represents the overall training pattern constraints following statistical learning theory \cite{Vapnik1999}.

Contextual constraints. According to Achille and Soatto \cite{Achille2018}, contextual constraints reveal through specific mechanisms that shape how information flows through the network. The cited paper discusses several possible contextual constraint mechanisms as sequential dependency or context bottleneck effect. We can write our constraint as

\begin{equation}
\delta = std\left( attention\_ weights\lbrack:windowsize\rbrack \right)
\end{equation}

Metric $std(attention\_ weights)$ reflects how consistently the model uses context (which is related to sequential dependency), shows how variable attention is distributed (which is related to the context bottleneck effect), and it's easy to compute during training. This measurement aligns with our theoretical framework, which postulates that contextual constraints emerge from the interaction between the model architecture (attention mechanism), training objectives (task sufficiency), and information bottleneck effects (layer stacking).

All three constraints directly affect the system's ability to explore its adjacent possible states following original TAP framework:

\begin{equation}
P(true|training) \propto (1 - \beta)(1 - \gamma)(1 - \delta)P(state|architecture)
\end{equation}

\subsection{Hypothesis 3: Path dependence influences problem-solving trajectories}

Our experimental design for analyzing path dependence focused on four complementary measurements, capturing different features of solution trajectory changes. As test data we used the high school mathematics subset of the MMLU dataset \cite{Hendrycks2020} with 30 questions to ensure statistical robustness. The four metrics are the following:

a) Building on trajectory analysis methods from stochastic processes \cite{Cornfeld2012}, we compared solution paths under normal and shuffled input conditions, measuring step count variations, consistency differences, and directness disparities. We used the expression

\begin{equation}
\mathrm{\Delta}_{path} = \left| M_{normal} - M_{shuffled} \right|
\end{equation}

where $M$ represents metrics including number of steps, path consistency, and solution directness. Solution paths were analyzed both in their original order and with shuffled choice presentations to assess path dependence effects.

b) Consistency through the model's hidden state representations is measured using cosine similarity between consecutive solution steps. We extracted the final layer's hidden states for each solution step and computed mean embeddings across the sequence length dimension. Last, we calculated cosine similarity between consecutive step embeddings. This approach follows methods established for analyzing neural network internal representations \cite{Elhage2021}. We used the following equation:

\begin{equation}
C = \frac{1}{n - 1}\sum_{i = 1}^{n - 1}{\cos{(s_{i},}}s_{i + 1})
\end{equation}

where $s_{i}$ represents the semantic embedding of step $i$ and $n$ is the total number of steps.

c) We quantified solution directness through a combination of step count and revision detection, where revisions were identified through specific linguistic markers (e.g., "actually", "instead", "correction") following approaches from solution path analysis \cite{Levine2022}. To compute the solution directness we use the following equation:

\begin{equation}
D = \frac{1}{1 + |steps|} + \frac{1}{1 + |revisions|}
\end{equation}

where $|steps|$ is the total number of solution steps, and $|revisions|$ counts backtracking instances.

d) Finally, the step length variations were computed with the following equation:

\begin{equation}
L_{diff} = L_{normal} - L_{shuffled}
\end{equation}

where $L$ is the average step length under normal and shuffled conditions.

This experimental design is rooted in existing methodologies for analyzing stochastic systems, which we adapted to the unique context of large language model evaluation. The approach aligns with previous work on analyzing non-ergodic behavior in complex systems \cite{Kauffman2019}. The key limitation of this setup is potential noise in path difference measurements for complex problems, addressed through multiple trials and robust statistical controls \cite{Goodfellow2016}. This approach enables systematic comparison of path dependence effects across different model architectures while maintaining statistical robustness and measurement reliability. Each metric captures a distinct aspect of path dependence, from direct trajectory differences to more subtle variations in solution characteristics. Following standard practices in language model evaluation, we used temperature sampling for generation.

\section{Experimental results}

\subsection{Phase Transitions in Semantic Space (H1)}

The first hypothesis proposes a correlation between the emergence of capabilities in language models and phase transitions in semantic space. The experimental results support this prediction, although there are significant differences in the occurrence of these transitions. These results may explain why models cannot perpetually expand their semantic complexity, as well as the predictable impact of resource allocation on model performance. Finally, it proves the inherent limitations of model scaling based on available resources.

\begin{figure}[htbp]
    \centering
    \includegraphics[width=0.75\textwidth]{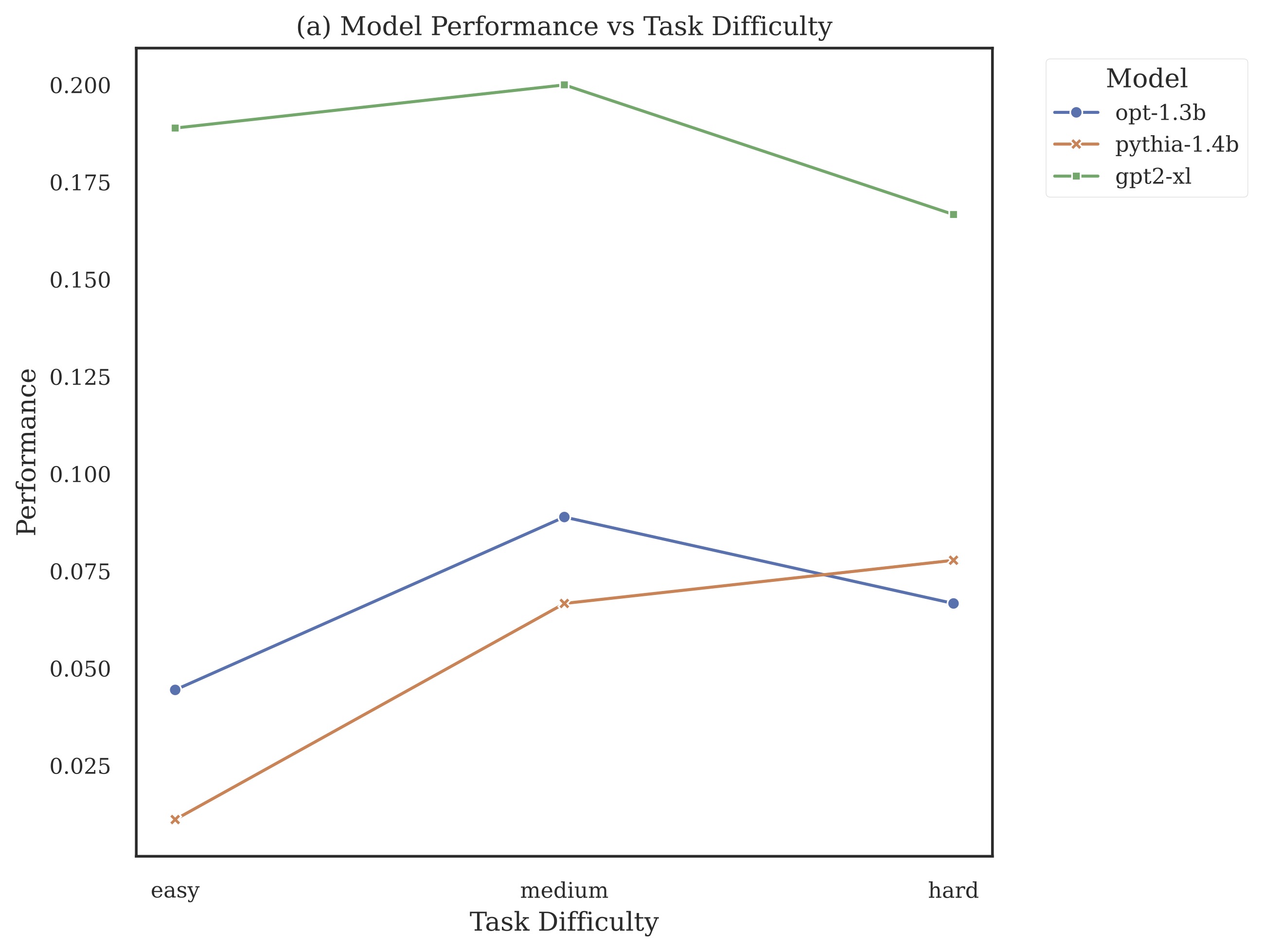}
    \caption{Model performance and task difficulty}
    \label{fig:model-performance}
\end{figure}

Figure \ref{fig:model-performance} illustrates how both gpt2-xl and opt-1.3B exhibit U-shaped inverted performance trajectories, with a peak of medium difficulty at different performance levels (gpt2-xl peak: 0.200, opt-1.3B peak: 0.089).  In contrast, pythia-1.4B shows a different pattern with a sharp initial increase from easy to medium, followed by continued improvement, unveiling a different form of phase transition in its capability space.

\begin{figure}[htbp]
    \centering
    \includegraphics[width=0.75\textwidth]{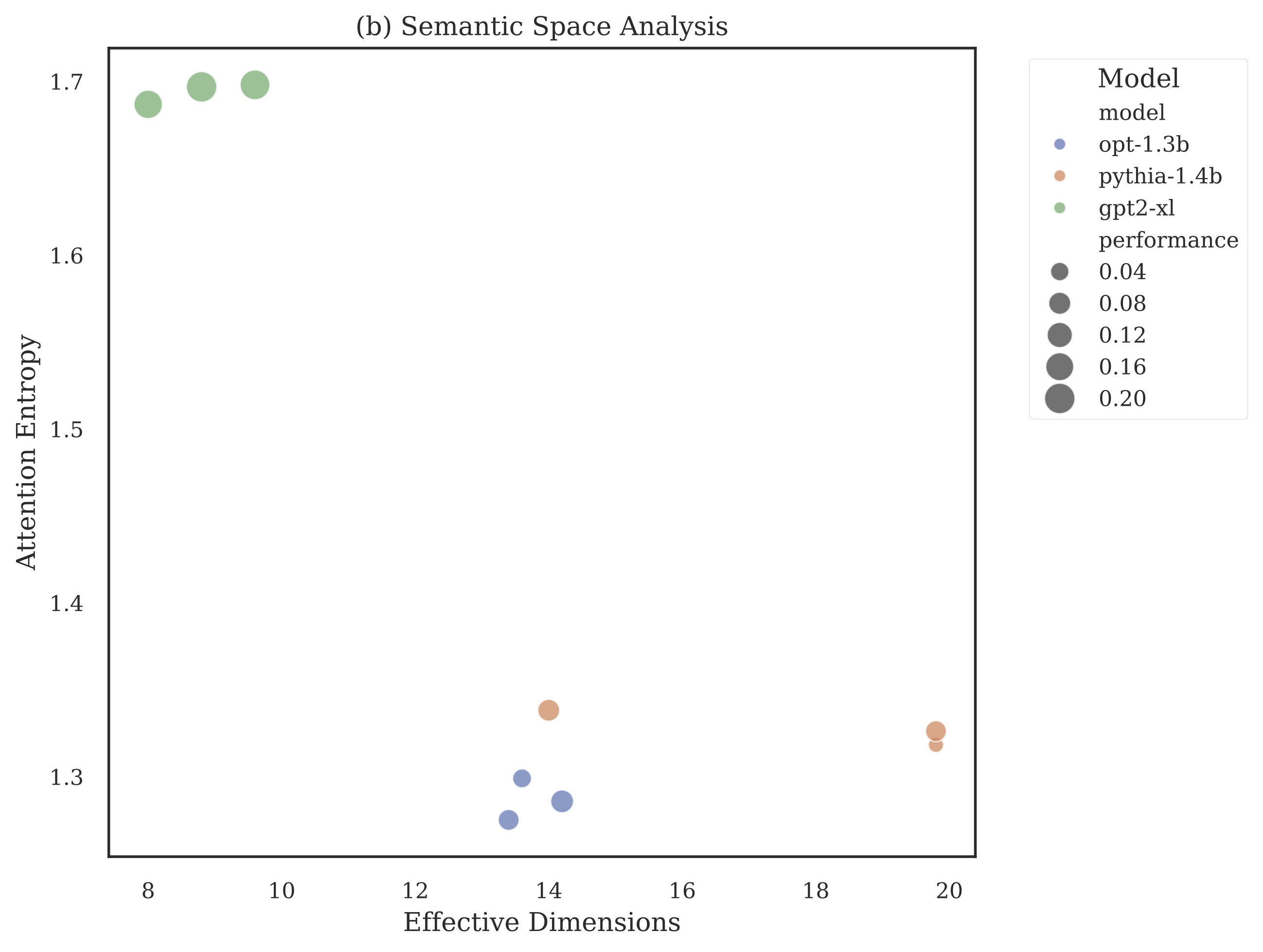}
    \caption{Semantic space analysis.}
    \label{fig:semantic-space}
\end{figure}

The semantic space analysis in Figure \ref{fig:semantic-space} shows how these transitions correlate with changes in the models' operational regime. gpt2-xl (green) operates in a distinctive regime characterized by a high attention entropy ($\sim$ 1.7) and a lower effective dimensionality (8-10 dimensions). This suggests that the model distributes attention broadly but in a more compact semantic space. In contrast, opt-1.3B (in orange) and pythia-1.4B (in blue) show lower attention entropy ($\sim$1.3-1.4) and higher effective dimensionality (14-20 dimensions). This could indicates more focused attention patterns but across a larger semantic space. The size of the dots represents model performance, providing additional insight into how these different regimes relate to model capabilities. This visualization reveals a fundamental trade-off in language model design: models can either operate with high entropy in a compact space (like gpt2-xl) or with more focused attention across a larger-dimensional space (like opt-1.3B and pythia-1.4B). These distinct operational regimes suggest different strategies for managing the complexity of language processing, with implications for how phase transitions in capabilities emerge in each architecture type.

\begin{figure}[htbp]
    \centering
    \includegraphics[width=0.75\textwidth]{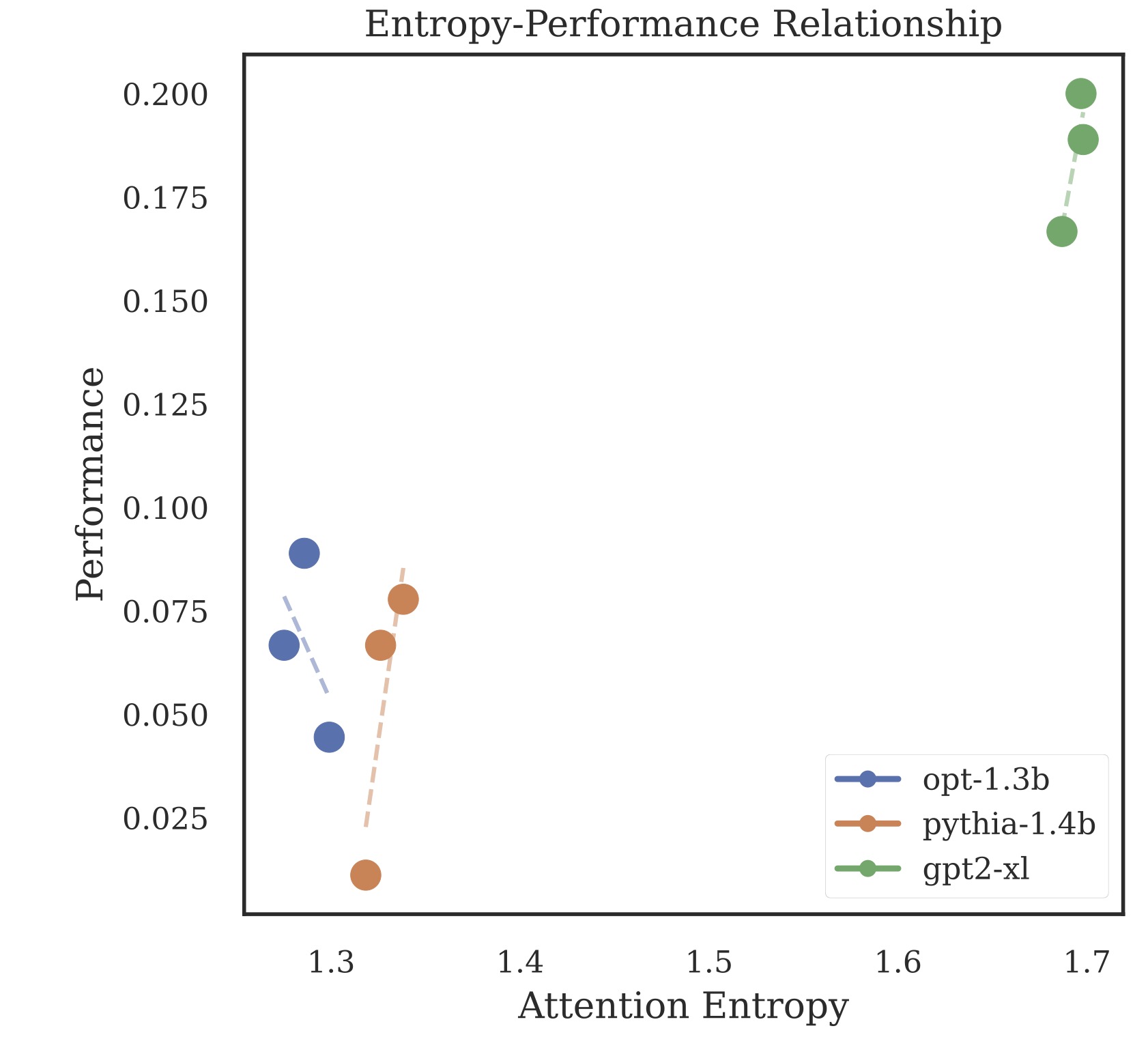}
    \caption{Correlation between performance and entropy.}
    \label{fig:performance-entropy}
\end{figure}

In Figure \ref{fig:performance-entropy}, strong correlations between performance and entropy for each model (gtp2-xl: 0.910, pythia-1.4B: 0.883) suggests that transitions follow organized patterns rather than random fluctuations. The varying stability metrics (gpt2-xl: 0.925, opt-1.3B: 0.728, pythia-1.4B: 0.438) further suggest that different models shows different types of phase transitions, from smooth progressions to sharp capability shifts.

\subsection{Constraint interactions shape capability boundaries (H2)}

Our second hypothesis consider that model capabilities are shaped by multiplicative interactions between architectural, training, and contextual constraints. The experimental results can be seen in the following tables.

Performance stability metrics shown in Table \ref{tab:performance-stability} suggest different behavior patterns across architectures. Model gpt2-xl shows higher overall performance (mean=2.3477) while maintaining moderate stability (CV=3.26\%). opt-1.3b shows the most stable performance (CV=2.68\%) but lower absolute performance (mean=1.8089), and pythia-1.4b has the highest variability (CV=6.97\%), suggesting less robust constraint management. The stability patterns indicate that architectural differences significantly influence how models manage constraint trade-offs, with gpt2-xl achieving the best balance between performance and stability.

\begin{table}[htbp]
\centering
\caption{Performance stability metrics.}
\label{tab:performance-stability}
\begin{tabular}{lcccccc}
\toprule
Model & Std & CV(\%) & Range & Min & Max & Mean \\
\midrule
op-1.3B & 0.0484 & 2.68 & 0.096 & 1.7639 & 1.8602 & 1.8089 \\
pythia-1.4B & 0.1281 & 6.97 & 0.2471 & 1.7340 & 1.9811 & 1.8380 \\
gpt2-xl & 0.0765 & 3.26 & 0.1461 & 2.2616 & 2.4077 & 2.3477 \\
\bottomrule
\end{tabular}
\end{table}

Table \ref{tab:constraints-analysis} shows the analysis of constraint interactions across different difficulty levels. Architectural constraints shows a systematic decrease with difficulty (from 6.2444 to 5.9117). Highest effect in easy tasks suggests more efficient architectural utilization, and significant drop in hard tasks point architectural strain. Training Constraints shows a non-monotonic behavior (from 0.1300 to 0.0000 and to 0.1261) with a valley at medium difficulty, suggesting a critical transition point. Recovery in hard tasks suggest an adaptive training dynamics. Contextual Constraints reveal a slight increase with difficulty level (from 0.0831 to 0.0935), being the most stable among all constraints. It suggests increasing reliance on context for complex tasks. Weighted constraints combination ("Performance" column in Table \ref{tab:constraints-analysis}) shows a peak performance in easy tasks (2.0829), then stabilizes around 1.94-1.97 for medium/hard tasks. These results provide an evidence that constraint interactions adapt dynamically to task complexity. The systematic decrease in architectural effects with increasing difficulty supports our hypothesis about adaptive constraint dynamics.

\begin{table}[htbp]
\centering
\caption{Analysis of constraints across difficulty levels. The performance column represents a weighted combination of architectural, training and contextual constraints effects (30\% $\beta$, 40\% $\gamma$, and 30\% $\delta$), plus an additional 10\% contribution from their interaction. $\beta$ are the architectural constraints, $\gamma$ training data constraints, and $\delta$ are contextual constraints.}
\label{tab:constraints-analysis}
\begin{tabular}{lcccc}
\toprule
 & $\beta$ & $\gamma$ & $\delta$ & Performance \\
\midrule
Easy & 6.2444 & 0.1300 & 0.0831 & 2.0829 \\
Medium & 6.2380 & 0.0000 & 0.0826 & 1.9414 \\
Hard & 5.9117 & 0.12611 & 0.0935 & 1.9702 \\
\bottomrule
\end{tabular}
\end{table}

\subsubsection{Constraints analysis}

Analysis of normalized constraint values in Table \ref{tab:normalized-constraints} shows several striking patterns. For example architectural constraints remain consistent across models (from 0.6348 to 0.6409), revealing a common architectural bottleneck. This suggests that despite different architectures, all three models hit similar fundamental limitations in how they can process information. Training constraints show higher variance (from 0.3333 to 0.5193), with pythia-1.4b model having the lowest normalized training constraint impact. This variation suggests that models learn and utilize training data differently. Contextual constraints shows a clear progression (from 0.3350 to 0.4054). The model gpt2-xl shows a bigger impact of context constraints, which is consistent with prior entropy analyses indicating a wider attention distribution for gpt2-xl.

\begin{table}[htbp]
\centering
\caption{Normalized constraints values. $\beta$ are the architectural constraints, $\gamma$ training data constraints, and $\delta$ are contextual constraints}
\label{tab:normalized-constraints}
\begin{tabular}{lccc}
\toprule
Model & $\beta$ & $\gamma$ & $\delta$ \\
\midrule
op-1.3B & 0.6348 & 0.5000 & 0.3350 \\
pythia-1.4B & 0.6409 & 0.3333 & 0.3562 \\
gpt2-xl & 0.6378 & 0.5193 & 0.4054 \\
\bottomrule
\end{tabular}
\end{table}

In Table \ref{tab:relative-constraints} we can see how training constraints consistently show the highest relative importance (from 0.7506 to 1.4952), pointing their primary role in shaping model's performance. Architectural constraints represent model-specific relative importance, with opt-1.3b (0.8325) and gpt2-xl (0.7314) showing similar dynamics. Contextual constraints shows lower but consistent relative importance (from 0.1581 to 0.5223). The high R$^2$ score values confirm the significance of these results.

\begin{table}[htbp]
\centering
\caption{Relative constraints importance. $\beta$ are the architectural constraints, $\gamma$ training data constraints, and $\delta$ are contextual constraints.}
\label{tab:relative-constraints}
\begin{tabular}{lcccr}
\toprule
Model & $\beta$ imp. & $\gamma$ imp. & $\delta$ imp. & R$^2$ score \\
\midrule
op-1.3B & 0.8325 & 1.4591 & 0.5223 & 0.9634 \\
pythia-1.4B & 0.1645 & 0.7506 & 0.1581 & 0.9984 \\
gpt2-xl & 0.7314 & 1.4952 & 0.1667 & 0.9513 \\
\bottomrule
\end{tabular}
\end{table}

\begin{table}[htbp]
\centering
\caption{Architectural constraints $\beta$, and training constraints $\gamma$ threshold.}
\label{tab:constraints-threshold}
\begin{tabular}{lcc}
\toprule
Model & $\beta$ threshold & $\gamma$ threshold \\
\midrule
opt-1.3B & 5.72 & 0.11 \\
pythia-1.4B & 5.80 & 0.11 \\
gpt2-xl & 7.45 & 0.06 \\
\bottomrule
\end{tabular}
\end{table}

Table \ref{tab:constraints-threshold} shows different thresholds in the effects of architectural and training constraints for all models. This supports our hypothesis that capability boundaries are influenced by the interactions of constraints. The gpt2-xl model has a higher architectural threshold in comparison to other models. This indicates improved performance, implying that architectural constraints significantly influence model capabilities.

\begin{figure}[htbp]
    \centering
    \includegraphics[width=0.75\textwidth]{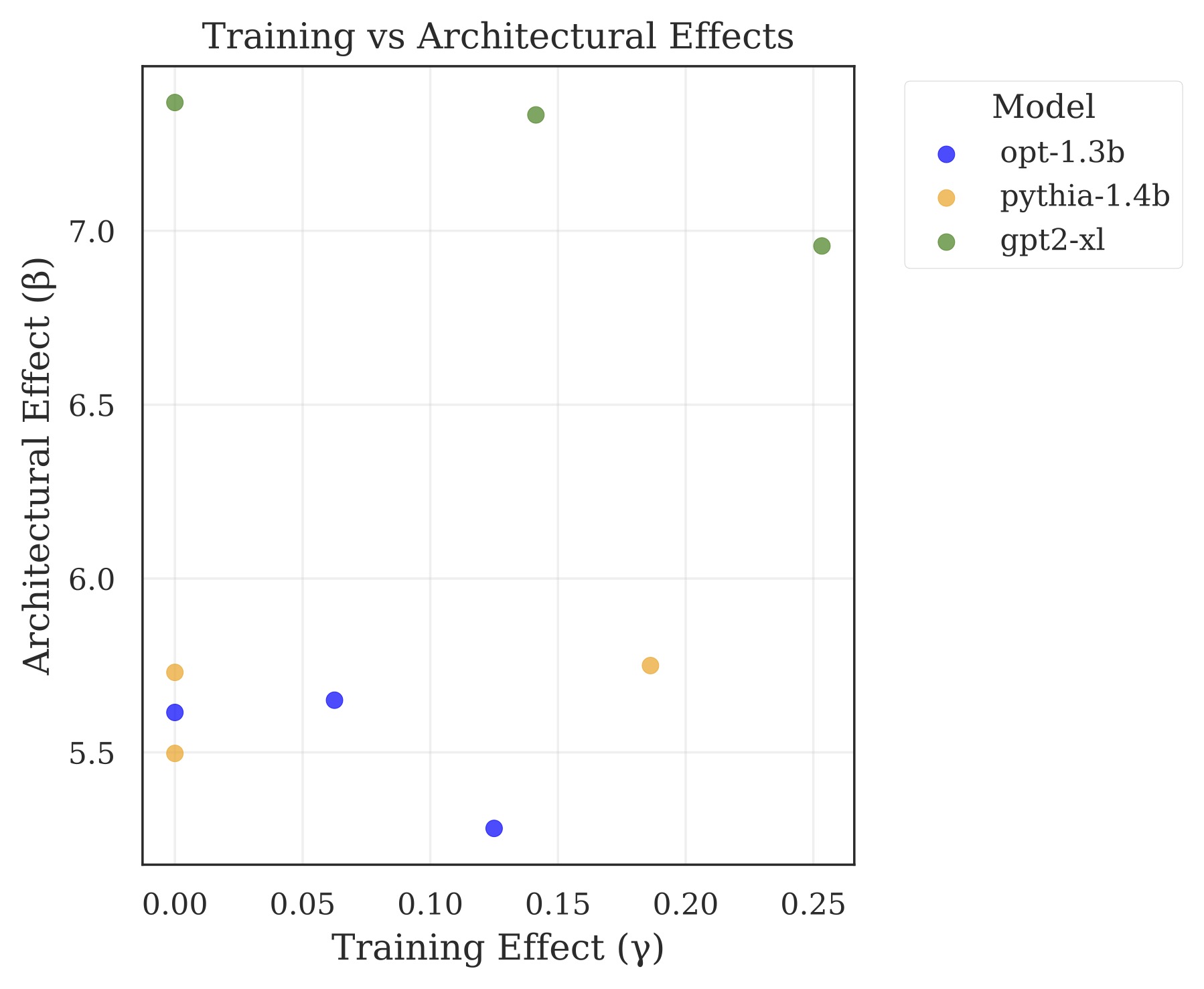}
    \caption{Training vs architectural constraints effects.}
    \label{fig:training-architectural}
\end{figure}

The correlation between training and architectural constraints shown in Figure \ref{fig:training-architectural} reveals uncorrelated dynamics. gpt2-xl (green dots) shows consistently high architectural effects ($\sim$7.0-7.5) regardless of training effects. opt-1.3B (blue dots) and pythia-1.4B (orange dots) cluster together at lower architectural effects ($\sim$5.5-6.0). This clear separation suggests fundamentally different operational regimes between gpt2-XL and the other models. Training effects range from 0.0 to 0.25 across all models with no clear pattern or correlation between training and architectural effects. Points are scattered horizontally, suggesting training effects vary independently of architectural constraints. The lack of correlation suggests architectural and training constraints operate independently. While gpt2-xl maintains higher architectural effects despite of training variations, opt-1.3B and pythia-1.4B show similar architectural behaviors across different training effects.

\begin{figure}[htbp]
    \centering
    \includegraphics[width=0.75\textwidth]{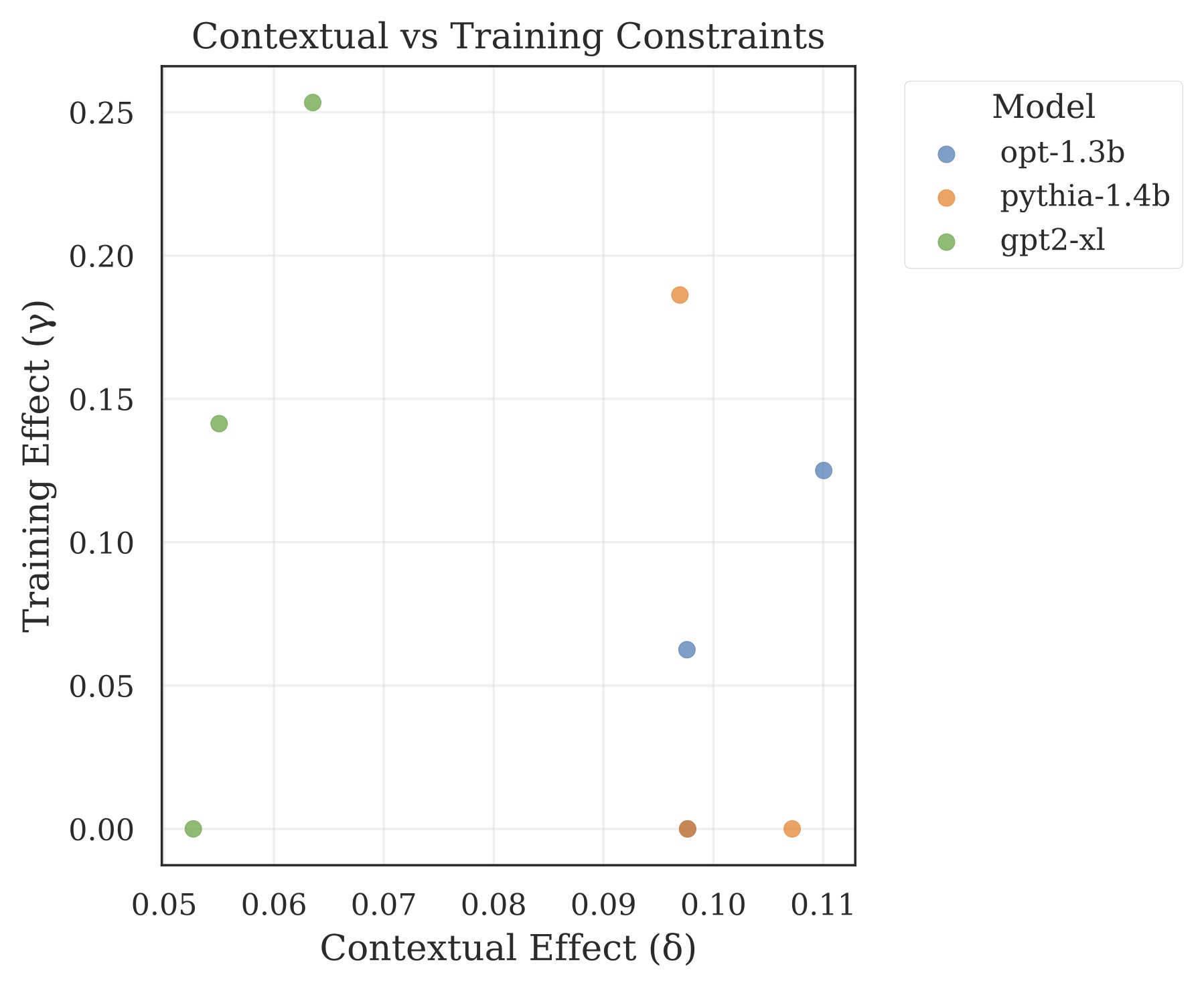}
    \caption{Training vs contextual constraints effects.}
    \label{fig:training-contextual}
\end{figure}

\begin{figure}[htbp]
    \centering
    \includegraphics[width=0.75\textwidth]{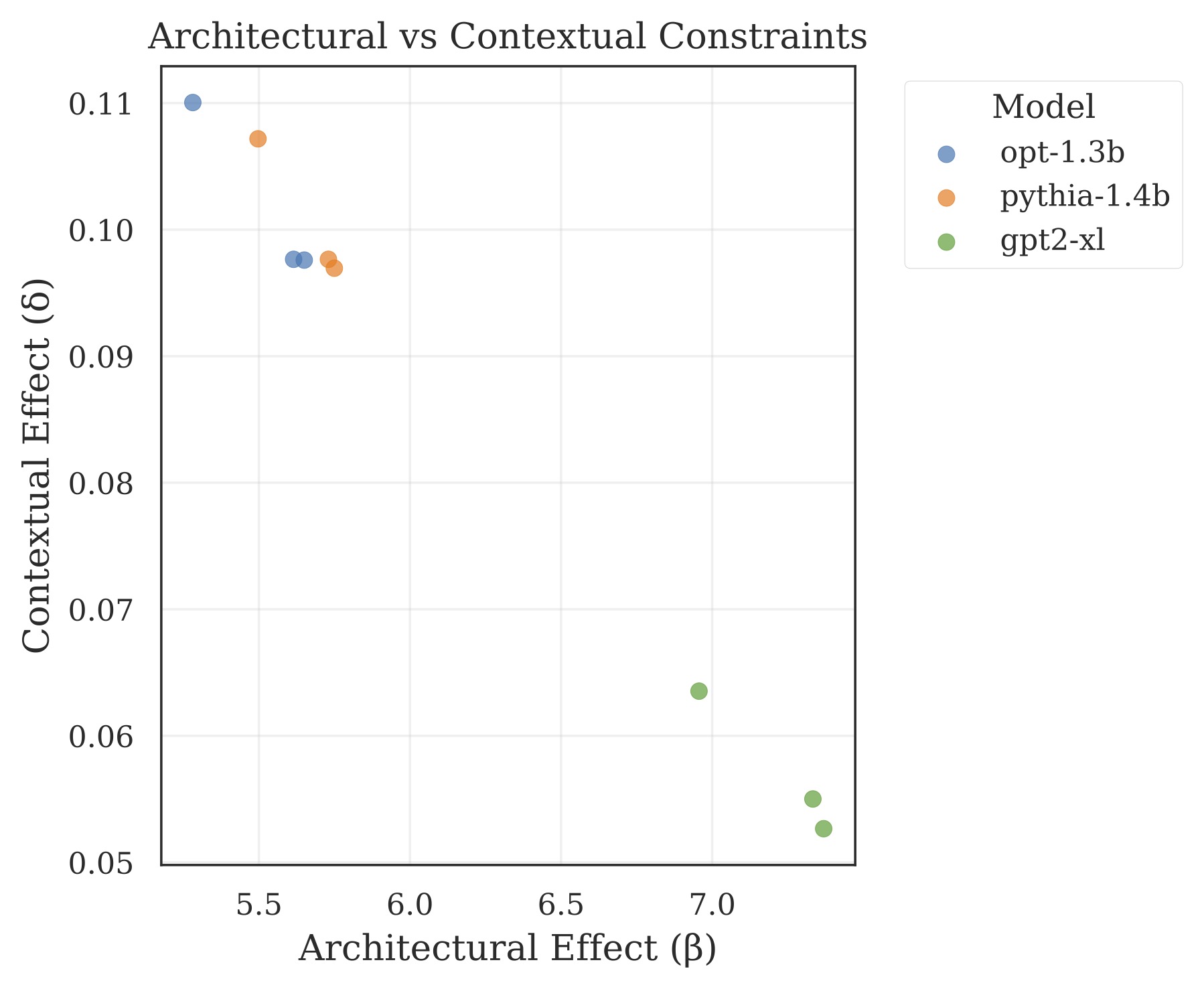}
    \caption{Architectural vs contextual constraints effects.}
    \label{fig:architectural-contextual}
\end{figure}

Training and contextual constraints correlations in Figure \ref{fig:training-contextual} vary significantly by model. Model gpt2-xl shows a precise positive correlation ($R^{2}$=1.0 with $p$-value=0.000), while pythia-1.4b and opt-1.3b models offers less significant correlation ($R^{2}$=-0.866 with $p$-value=0.333, and $R^{2}$=0.500 with $p$-value=0.667 respectively). Figure \ref{fig:architectural-contextual} illustrates how architectural and contextual constraints correlations show precise negative correlation across all models ($R^{2}$=-1.0 with $p$-value=0.000), indicating a fundamental trade-off between both constraints.

\begin{figure}[htbp]
    \centering
    \includegraphics[width=0.75\textwidth]{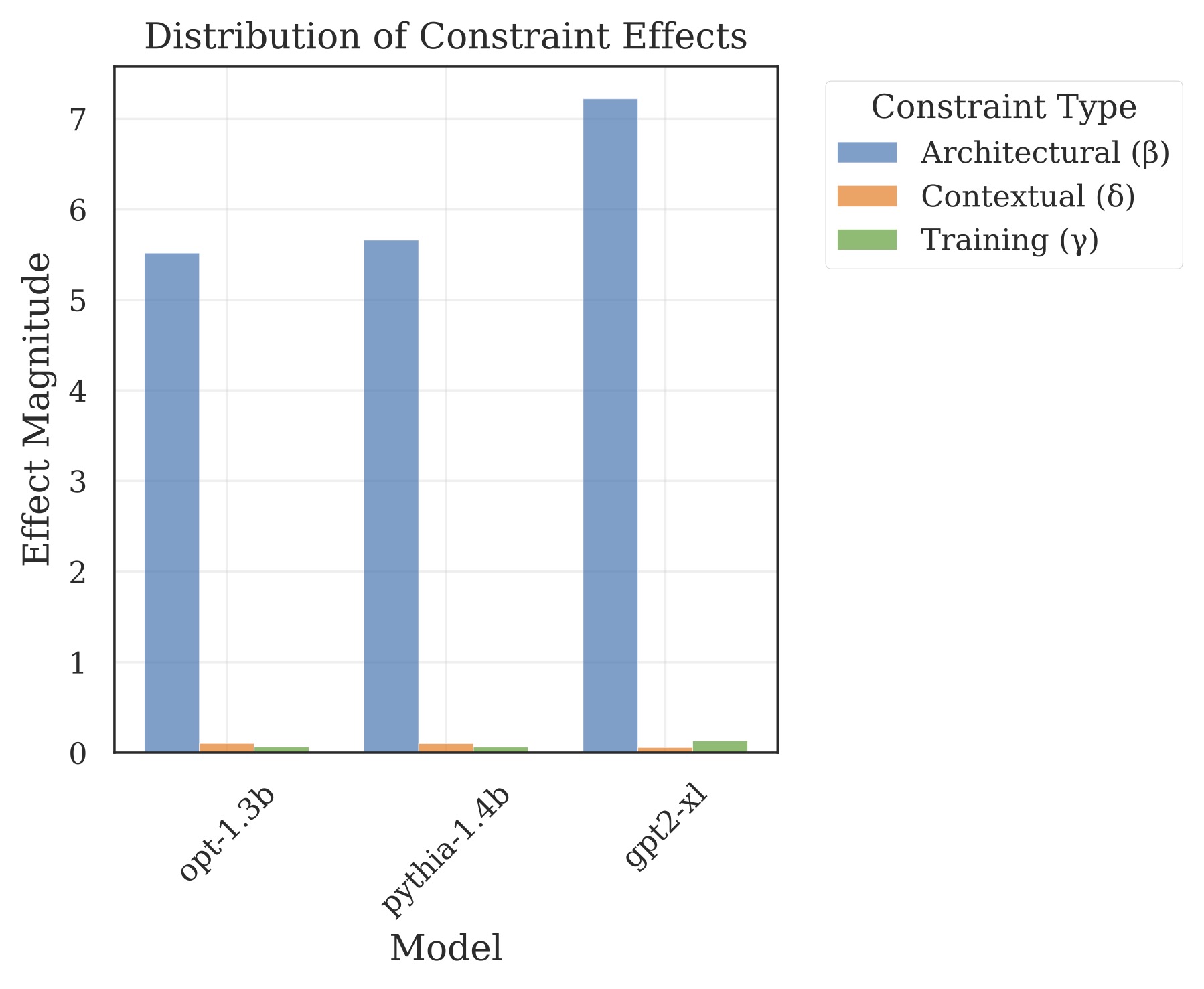}
    \caption{Constraints distribution for different models.}
    \label{fig:constraints-distribution}
\end{figure}

Figure \ref{fig:constraints-distribution} shows an interesting distribution of constraint effects across three different models, and its results connect directly to our theoretical framework in several important ways. The influence of architectural constraints in all three models compared to contextual and training constraints (near $0$) aligns with our theoretical prediction that architectural constraints are fixed at initialization and fundamentally limit the model's capability space. Particularly interesting is that gpt2-xl shows the highest architectural constraint, suggesting its architecture more strictly bounds its possibility space. These results reflect our theoretical framework's emphasis on architectural constraints through $R\left( C_{t} \right)$. However, we need additional independent evidence to validate whether this emphasis accurately represents the real dynamics of language models.

\begin{figure}[htbp]
    \centering
    \includegraphics[width=0.75\textwidth]{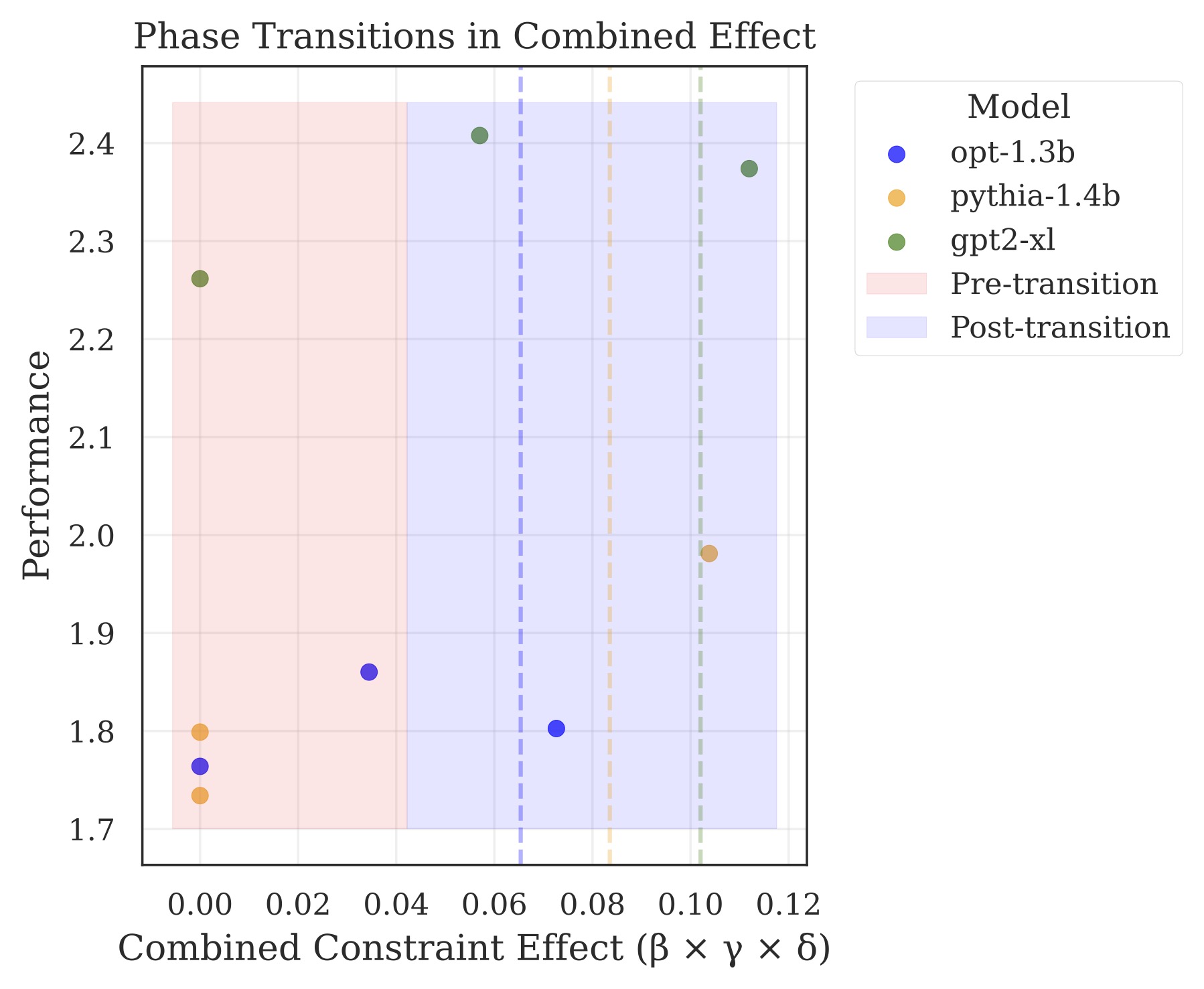}
    \caption{Phase transitions for combined constraints across different models.}
    \label{fig:phase-transitions}
\end{figure}

Figure \ref{fig:phase-transitions} shows the relationship between combined constraint effects ($\beta \times \gamma \times \delta$) and model performance, revealing critical phase transitions in the constraint space. The plot demonstrates distinct model behaviors and clear transition points. Model gpt2-xl has the highest performance (from 2.2 to 2.4) maintaining a stable performance across a wider range of combined constraint values (from 0.06 to 0.12). In contrast, opt-1.3b and pythia-1.4b models operate in a lower performance region (from 1.7 to 2.0) and show more abrupt transitions. Each model demonstrates a distinct critical threshold (marked with vertical dashed lines): gpt2-xl at 0.10, pythia-1.4b at 0.08, and opt-1.3b at 0.07. The pre-transition region (red shaded area) represents a phase where models operate below their optimal constraint balance, while the post-transition region (blue shaded area) indicates where models achieve better constraint integration. These thresholds mark points where models transition from lower to higher performance states. Higher thresholds correlate with better overall performance. For example model gpt2-xl not only has the highest threshold but also maintains more consistent performance in the post-transition region. This suggests that its architecture achieves a more robust balance of constraints. The sharp performance shifts at these thresholds indicate the non-linear nature of constraint interactions and the existence of critical points where model behavior has a symmetry breaking.

Our experimental results provide strong support for hypothesis 2 through various evidences. First, the performance-stability analysis shows how different architectures manage constraints distinctly, with gpt2-xl achieving optimal balance while maintaining high performance. Second, the constraint interaction analysis across difficulty levels reveals dynamic adaptation, evidenced by the systematic decrease in architectural effects and the architectural/contextual constraints ratio demonstrating how constraints adapts to task complexity. Third, the correlation analysis reveals fundamental trade-offs, particularly the perfect negative correlation between architectural and contextual constraints. Finally, the phase transition analysis reveals clear threshold effects in both architectural impact and combined constraint interactions, where each model shows distinct but related critical points. These transitions and model-specific thresholds demonstrate that performance emerges from multiplicative interactions between constraints rather than simple additive effects, conclusively supporting our hypothesis that model capabilities are shaped by complex interactions between architectural, training), and contextual constraints.

\subsection{Constraint interactions shape capability boundaries (H3)}

Results for Hypothesis 3 can be seen in the following figures. Figure \ref{fig:path-differences} shows the path differences across models. Model opt-1.3b shows the highest number of step differences (0.447), suggesting strongest path dependence in solution approach. Models pythia-1.4b and gpt2-xl show moderate step differences (0.347 and 0.287 respectively), suggesting more stable solution strategies despite input order.

\begin{figure}[htbp]
    \centering
    \includegraphics[width=0.75\textwidth]{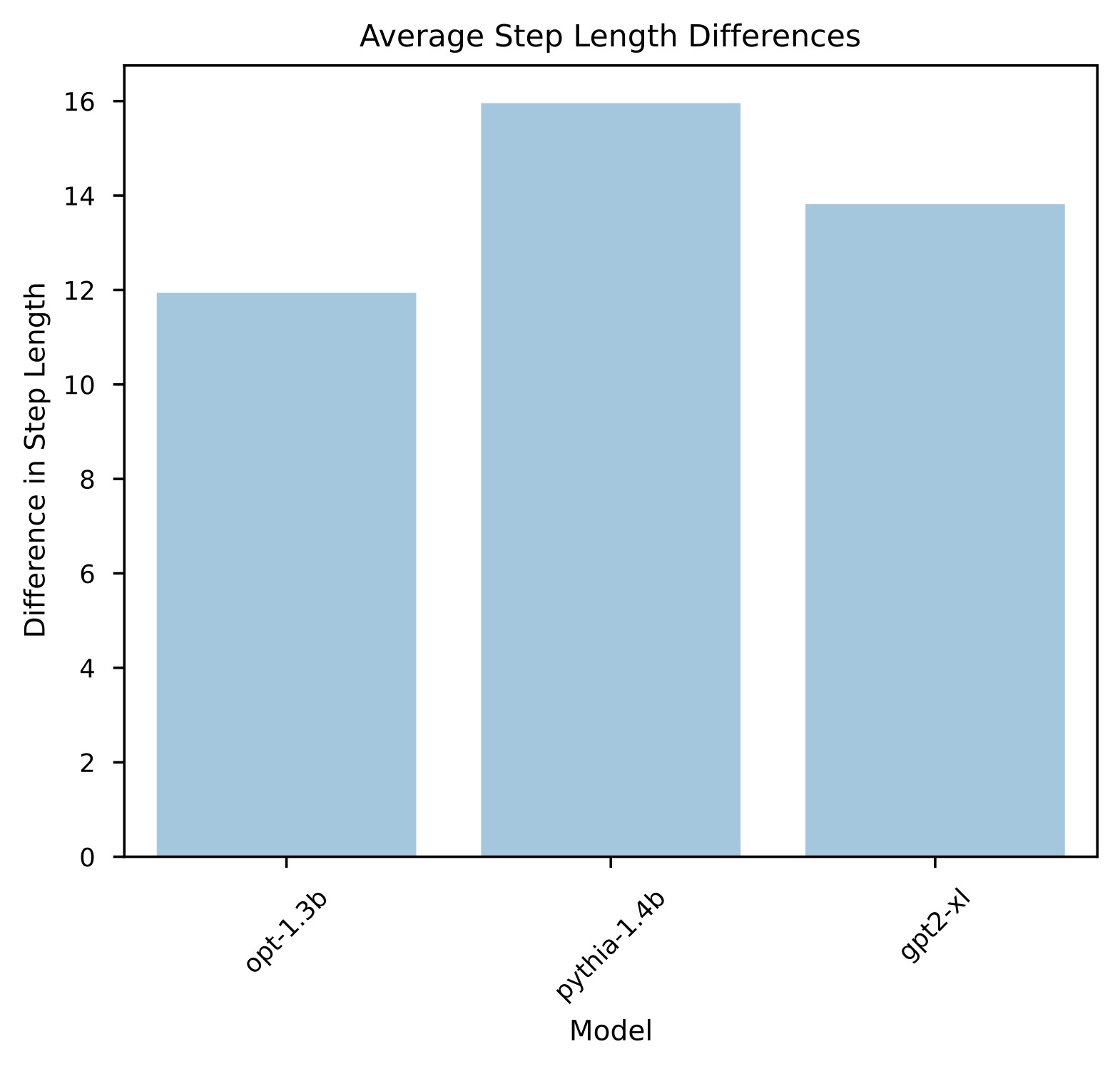}
    \caption{Step length analysis across models.}
    \label{fig:step-length}
\end{figure}

Figure \ref{fig:step-length} compares the step length variability across models. Pythia-1.4b shows the largest step length difference (15.957), indicating high sensitivity to input ordering in terms of solution verbosity. Model opt-1.3b shows moderate step length variation (11.941) and model gpt2-xl shows high but consistent step length differences (13.819). Figure \ref{fig:step-length} shows the consistency vs directness analysis dynamics not revealing any important pattern to be considered.

\begin{figure}[htbp]
    \centering
    \includegraphics[width=0.75\textwidth]{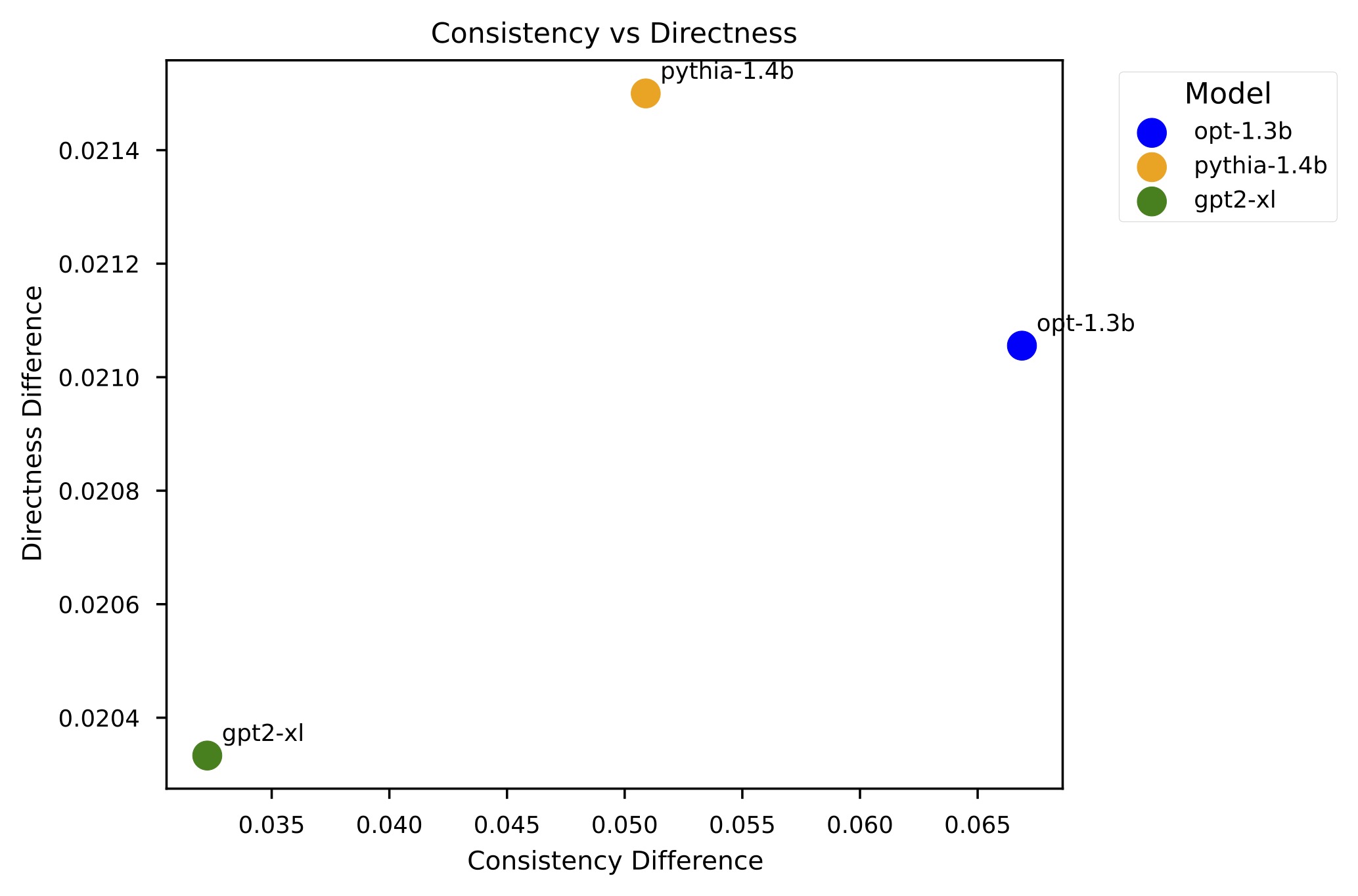}
    \caption{Path differences across models}
    \label{fig:path-differences-2}
\end{figure}

In Figure \ref{fig:path-differences-2} we observe that all models has similar directness differences (0.020-0.021), suggesting comparable efficiency in reaching solutions regardless of path. Consistency differences vary more significantly with model opt-1.3b having the highest value (0.067) and gpt2-xl the lowest (0.032). The tight clustering of directness differences despite varying consistency differences suggests that models maintain solution efficiency even when following different paths.

\begin{figure}[htbp]
    \centering
    \includegraphics[width=0.75\textwidth]{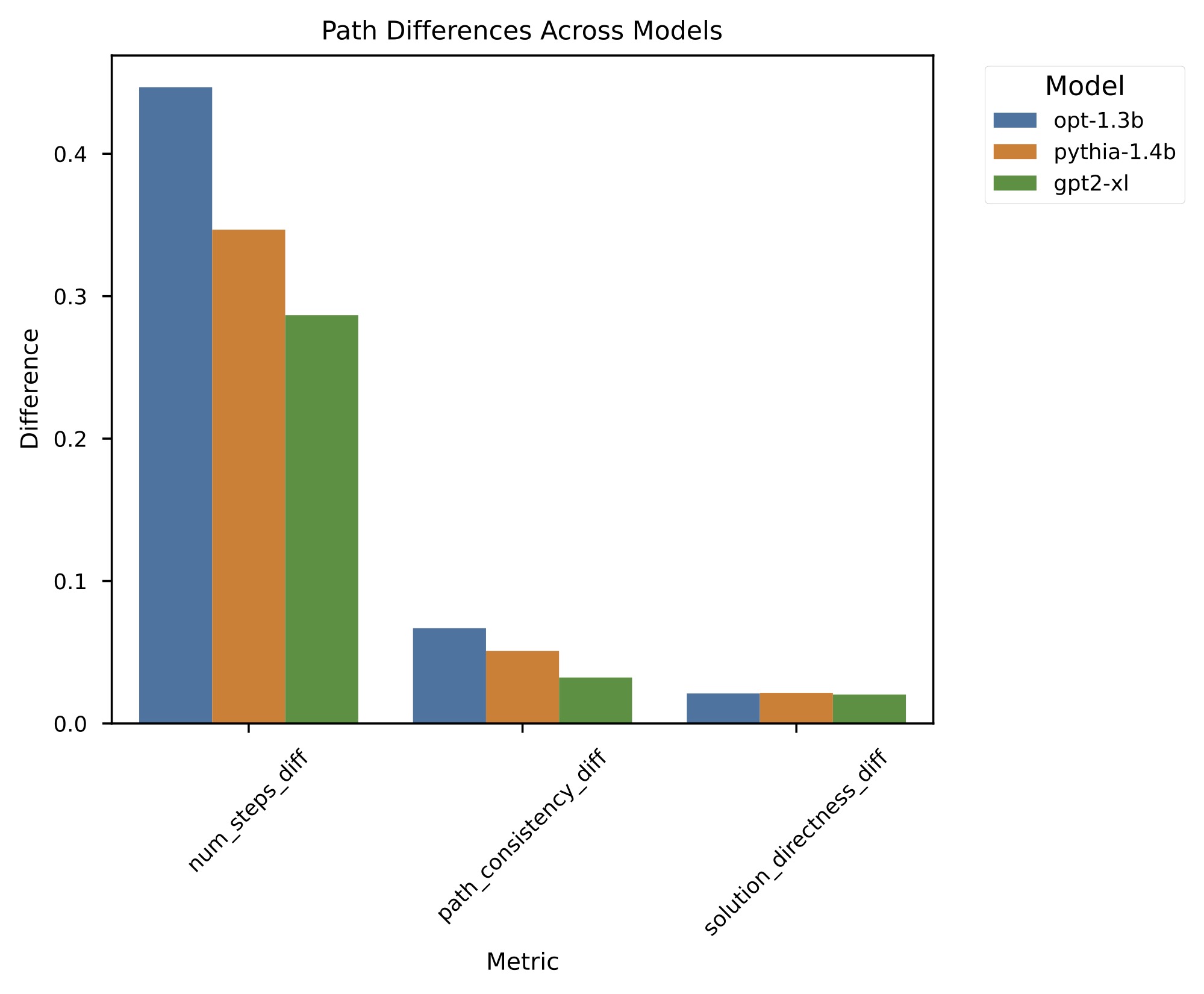}
    \caption{Path differences across models.}
    \label{fig:path-differences}
\end{figure}

\begin{figure}[htbp]
    \centering
    \includegraphics[width=0.75\textwidth]{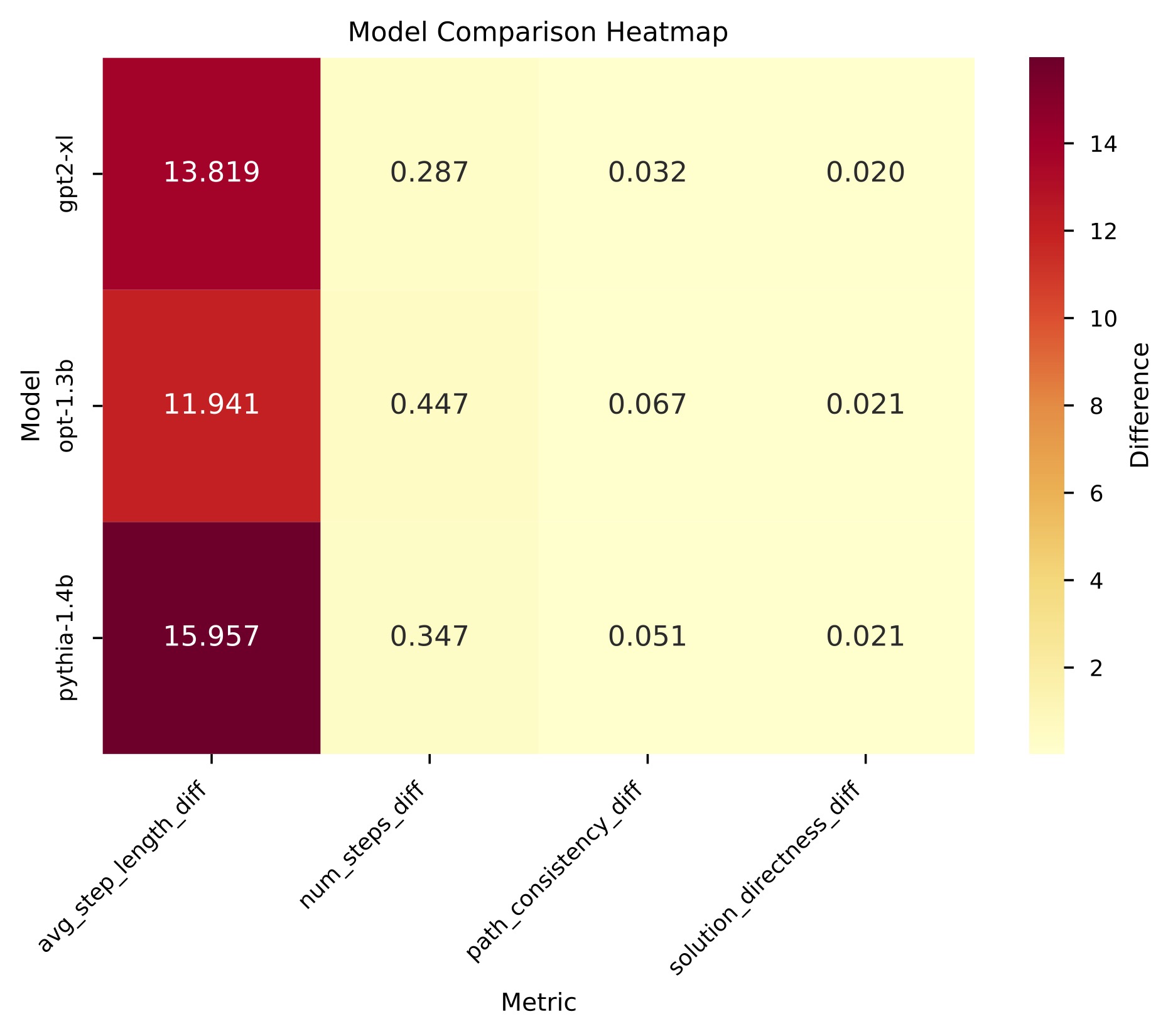}
    \caption{Path differences across models.}
    \label{fig:path-differences-3}
\end{figure}

Figure \ref{fig:path-differences-3} shows a comparison of different characteristics across the models. The model gpt2-xl suggests the most balanced performance across different metrics, showing moderate differences in step length and the lowest variation in consistency. The opt-1.3b model demonstrates significant path dependence while ensuring consistent solution directness. Model pythia-1.4b has the greatest variability in step length while also showing moderate levels of consistency and directness metrics.

These results provide support for H3, suggesting that path dependence significantly affects problem-solving trajectories across all models. While all models exhibit path dependence, they use different strategies to maintain solution quality: gpt2-xl prioritizes consistency, opt-1.3b balances between path exploration and directness, and pythia-1.4b shows high adaptability in solution length while maintaining solution quality. The observed variations in path-dependent strategies align with our theoretical framework's predictions about non-ergodic exploration in constrained possibility spaces. gpt2-xl's higher consistency (lowest consistency difference 0.032) reflects its architectural capacity for parallel processing through multiple attention heads \cite{Vaswani2017}, enabling stable representation maintenance across different solution paths. Model opt-1.3B's balanced approach, with moderate step differences (0.447), emerges from its enhanced layer connectivity that facilitates diversified path exploration while maintaining solution coherence \cite{Zhang2019}. Pythia-1.4B's shows high adaptability in solution length (15.957) while maintaining directness (0.021) indicating that position-aware processing through scaled rotary embeddings enables flexible path exploration within semantic constraints \cite{Biderman2023}.

These architectural differences manifest in our TAP equation through the hierarchical function $g_{l}$, where each model's specific attention mechanisms and layer connectivity patterns create distinct mappings between token combinations and semantic space. The varying magnitudes of path differences across models quantify how architectural constraints shape the adjacent possible states accessible during problem-solving, supporting our framework's prediction that capability emergence follows architecture-dependent trajectories through the possibility space.

\section{Discussion}

Our experimental results provide strong support for the application of TAP theory to understanding emergent capabilities in large language models, while also revealing important nuances in how these systems navigate their possibility spaces. The findings validate our proposed resource-bounded TAP equation and demonstrate that language models exhibit behavior patterns consistent with complex biological systems, particularly in terms of phase transitions, constraint interactions, and path dependence.

\subsection{Theoretical framework validation}

The experimental results provide substantial support for our theoretical framework across several important areas.

First, the observation of clear phase transitions in semantic space aligns with TAP theory's prediction of discrete shifts in capability as systems explore their adjacent possible states. The inverted U-shaped performance curves observed in gpt2-xl and opt-1.3B suggest that these transitions are not simply cumulative improvements but rather represent fundamental reorganizations of the models' operational regimes. This behavior parallels Kauffman's description of biological systems transitioning between distinct organizational states.

Second, the multiplicative interaction of constraints observed in our experiments supports our theoretical decision to model constraint interactions multiplicatively instead of additively in our TAP equation. The observed negative correlation between architectural and contextual constraints indicates that these constraints operate as fundamentally interconnected parameters, similar to the combined constraints found in biological systems within metabolic networks.

Third, the strong path dependence observed across all models confirms the non-ergodic nature of these systems, a key assumption in our theoretical framework. The varying consistency differences between models while maintaining similar directness metrics suggests that, like in biological systems, language models explore their possibility spaces through restricted but efficient pathways.

\subsection{Emergence and phase transitions}

Recent empirical research has provided specific evidence for emergence patterns in language models. Wei et al. (2022) documented discontinuous improvements across 23 different capabilities in PaLM, finding that abilities like multi-step reasoning emerged suddenly at certain model scales rather than improving gradually. Analogously, Ganguli et al. (2022) analyzed the predictability of model capabilities during scaling, differentiating between gradual improvements and unexpected behavioral changes. Recent work by \cite{Schaeffer2024} explored the nature of emergent abilities, developing more rigorous methods for categorizing between basic emergence and scaling effects. These works provide actual evidence for non-random patterns in capability emergence, setting the stage for our analysis of the diverse behavior patterns observed across models.

In our research, the diverse behavior patterns observed across models - from gpt2-xl's balanced performance to pythia-1.4B's continued improvement pattern - suggest that emergence can appear through different mechanisms depending on architectural choices and constraint distributions. This observation aligns with Holland (2006) characterization of emergence in complex adaptive systems, where global patterns arise from local interactions under varying constraint conditions. The correlation between performance and entropy suggests that these transitions follow organized patterns rather than random fluctuations, relating to Prigogine's work on dissipative structures where order emerges from the interaction between system dynamics and environmental constraints \cite{Prigogine2018}. The systematized nature of transitions we observed supports our theoretical framework's prediction that capability emergence follows constrained exploration paths rather than random search. This aligns with Haken's synergetics theory \cite{Haken1993}, which describes how collective behavior emerges through self-organization under constraints. Our findings follows the theory of self-organized criticality systems \cite{Bak2013, Kauffman1993, Markovic2014}, as models appear to naturally evolve toward critical states where new capabilities emerge. The architectural dependency of emergence patterns confirm a hierarchical organization in complex systems, where different architectural configurations lead to distinct emergent properties \cite{Simon2012}. This systematized path-dependent nature found in emergence in language models provides a connection between deterministic phase transitions in physical systems and the more complex emergence patterns seen in biological systems, suggesting a new category of emergent phenomena in artificial intelligence systems that require further theoretical investigation.

While previous research proved the existence of emergent capabilities in language models, our work provides a theoretical framework that explains why and how these capabilities emerge through using non-ergodic dynamics and the adjacent possible theory. By proving that language models operate as non-ergodic systems and demonstrating how their capability emergence is shaped by multiplicative constraint interactions, we move beyond descriptive observations to a mechanistic understanding of emergence.

\subsection{Constraint dynamics, path dependence and non-ergodicity}

The varying stability metrics across models suggest that different architectures create distinct hierarchical organizations of constraints, affecting how capabilities emerge The systematic decrease in architectural constraints with increasing task difficulty indicates that models adaptively redistribute their computational resources as tasks become more complex, supporting our resource-bounded formulation of the TAP equation. The non-monotonic behavior of training suggests that learned patterns play a complex role in shaping the adjacent possible space, similar to how biological systems' past adaptations influence their future possibilities.

The path dependence results provide strong evidence for the non-ergodic nature of language models, but with important qualifications. The variation in step differences across models suggests that architectural choices can significantly influence the degree of path dependence. This finding has important implications for our theoretical framework, suggesting that while all models operate in non-ergodic regimes, the strength of historical dependence can be modulated through architectural design. This parallels biological systems where different organizational structures can lead to varying degrees of historical contingency.

We have seen that models used in our experiments adopts distinct strategies to maintain solution quality. This strategies could be related to their architectures. For example, gpt2-xl's consistency priority can be related to its larger number of attention heads (25 heads per layer) and relatively smaller head dimension, enabling better parallel processing of information \cite{Radford2019}. This architectural choice explains its observed preference for consistency, as multiple attention heads can maintain stable representations across different solution paths.

Model opt-1.3B's balanced approach can be related to model's additional skip connections and modified layer normalization positions compared to standard transformer architectures \cite{Zhang2022}. These architectural features facilitate information flow between different layers, explaining its balanced approach between path exploration and directness. The higher step differences and maintained directness metrics suggest that the enhanced layer connectivity enables the model to explore different paths while maintaining solution coherence.

Pythia's architecture incorporates scaled rotary embeddings and modified attention patterns that enhance its position-aware processing \cite{Biderman2023}. This architectural choice explains its observed high adaptability in solution length while maintaining quality. The largest step length differences and consistent solution quality metrics align with its enhanced position-aware processing capabilities.

These architectural-behavioral correlations suggest a generalizable principle: the distribution and structure of attention mechanisms fundamentally shape how models navigate their possibility spaces. Models with more parallel processing capabilities tend toward consistency-focused strategies, while those with enhanced layer connectivity or position-aware processing enable more flexible exploration strategies.

This finding has significant implications for our theoretical framework, suggesting that although all models operate in non-ergodic regimes, both the extent and type of historical dependence can be adapted through architectural design. This is analogous to biological systems, where diverse organizational structures may result in differing levels of historical contingency. The capacity of predicting path dependence strategies based on architectural features indicates that our TAP framework captures the fundamental principles about the influence of system structure on the exploration of possibility space.

\section{Practical implications and future directions}

Our theoretical framework and experimental findings could have significant implications for AI development while suggesting important directions for future research. In the domain of model explainability, the TAP framework provides novel tools for understanding model behavior through constraint interactions and phase transitions. Recent research on emergence in large language models \cite{Arora2023, Chen2024, Schaeffer2024, Wei2022} suggests that capability shifts can be predicted through careful monitoring of model behavior, aligning with our observations of clear thresholds in capability emergence. These insights complement recent advances in mechanistic interpretability \cite{Bereska2024, Conmy2023, Liu2023, Rai2024} and neural network interpretation \cite{Elhage2021, Elhage2022}.

The resource-bounded TAP equation (Equation 34) provides specific guidance for designing new model architectures. The multiplicative nature of constraint interactions suggests that architectural improvements should focus on balanced enhancement of all constraints rather than optimizing single components. This understanding leads to several important architectural innovations. Attention mechanisms could be designed to maintain consistent entropy across different operational regimes, while layer connectivity patterns could facilitate both information preservation and flexible path exploration. Position-aware processing could be integrated throughout the architecture to enable adaptive context utilization. These principles extend to dynamic routing mechanisms that enable flexible path exploration and adaptive connectivity patterns supporting multiple solution trajectories.

The framework has particular relevance for AI alignment, where new research reports the importance of understanding how models internalize training objectives \cite{Askell2021}. The non-ergodic nature of language models suggests that alignment strategies must account for path dependence, while the multiplicative nature of constraint interactions indicates that controlling multiple constraints simultaneously might be more effective than focusing on individual ones. This aligns with recent theoretical work on multi-constraint optimization in AI systems \cite{DoshiVelez2017} and phase-aware training methods \cite{Nanda2023}.

Our framework enables deliberate design for specific phase transitions in model development. Architectural features can be tuned to target desired capability emergence points, while resource allocation can be optimized based on predicted transition thresholds. Critical points in semantic space expansion can be engineered through careful constraint balance. This approach suggests that AGI capabilities might emerge through discrete shifts rather than continuous improvement. Our findings indicate that a intentional strategy for specific phase transitions in AGI development might bring greater advantages than solely depending on scaling approaches. New research on architectural innovation \cite{Arnold2024} and compute-optimal scaling \cite{Alabdulmohsin2024} supports this insight.

The theoretical framework aligns with LeCun (2022) approach to autonomous machine intelligence. New implementations of H-JEPA \cite{Assran2023} test how hierarchical organizations might naturally emerge from constraint interactions. The observed path dependence in problem-solving strategies suggests that world model formation follows similar constrained exploration patterns, supporting recent advances in predictive modeling \cite{Ha2018, Matsuo2022}.

Likely, several promising research directions emerge from our work. The theoretical framework can be extended to develop more precise mathematical models of constraint interaction dynamics and investigate relationships between phase transitions and optimization landscapes. Practical applications include the development of real-time analysis tools based on the TAP framework and the implementation of phase-aware training algorithms. The framework also suggests rich opportunities for interdisciplinary research, particularly in exploring parallels between AI and biological learning systems through the TAP lens.

Resource management in future architectures will require built-in phase transition monitoring capabilities and adaptive allocation based on capability emergence patterns. These systems should incorporate flexible computational pathways supporting diverse problem-solving strategies while maintaining balanced resource utilization. Such architectural innovations, guided by our theoretical framework, could lead to more efficient and capable language models that exhibit more controlled and predictable emergence of capabilities.

This work opens new avenues for research while providing practical tools for immediate application in AI development. The integration of constraint-aware design principles with phase transition engineering and non-ergodic architecture principles offers a comprehensive approach to advancing AI systems. Through careful application of these principles, we can work toward developing more robust, interpretable, and capable AI systems that exhibit predictable and controllable emergence of capabilities.

Our proposal in this paper is based on Stuart Kauffman's theory of the adjacent possible. Kauffman introduced other ideas that could be very useful in the field of artificial intelligence. Future research could explore how autonomous agents in language models maximize their average rate of exploration of adjacent possible states while preserving coherent functionality, similar to capability expansion in biological systems. A deeper understanding of molecular autonomous agents' principles could provide new insights into how language models balance exploration of novel states with maintenance of existing capabilities, particularly in continual learning scenarios. Kauffman's NK fitness landscape model \cite{Kauffman2000} could provide a robust framework for understanding how language models navigate their possibility spaces through the interaction of N components (architectural elements, attention mechanisms, layer connectivity) and K epistatic interactions (constraint relationships). Future research could further explore how the roughness of these landscapes, defined by the degree of interdependence between components \cite{Kauffman1987}, influences the emergence of capabilities and the stability of model behavior. This perspective suggests studying how different architectural choices create varying degrees of landscape ruggedness, potentially explaining why some models exhibit more robust capability emergence while others show fragility or unpredictability. Adaptive walks on correlated fitness landscapes \cite{Kauffman1987} could further provide novel training strategies that effectively balance exploration of promising regions with exploitation of established solutions.

\bibliographystyle{plainnat}

\end{document}